\documentclass[lettersize,journal]{IEEEtran}
\usepackage{amsmath,amsfonts}
\usepackage{algorithmic}
\usepackage{algorithm}
\usepackage{array}
\usepackage[caption=false,font=normalsize,labelfont=sf,textfont=sf]{subfig}
\usepackage{textcomp}
\usepackage{stfloats}
\usepackage{url}
\usepackage{verbatim}
\usepackage{graphicx}
\usepackage{cite}
\usepackage{multirow}
\usepackage{caption}
\usepackage{tabularray}
\usepackage{array}
\usepackage{graphbox}
\newcommand{\se}{$\mathtt{SE}$ }
\newcommand{\secompact}{$\mathtt{SE}$}

\begin{document}

\title{Differential Morphological Profile Neural Networks\\for Semantic Segmentation}

\author{David Huangal,~\IEEEmembership{Student Member,~IEEE}, J. Alex Hurt,~\IEEEmembership{Member,~IEEE}}

\maketitle

\begin{abstract}
Semantic segmentation of overhead remote sensing imagery enables applications in mapping, urban planning, and disaster response. State-of-the-art segmentation networks are typically developed and tuned on ground-perspective photographs and do not directly address remote sensing challenges such as extreme scale variation, foreground-background imbalance, and large image sizes.
We explore the incorporation of the differential morphological profile (DMP), a multi-scale shape extraction method based on grayscale morphology, into modern segmentation networks.
Prior studies have shown that the DMP can provide critical shape information to Deep Neural Networks to enable superior detection and classification performance in overhead imagery.
In this work, we extend prior DMPNet work beyond classification and object detection by integrating DMP features into three state-of-the-art convolutional and transformer semantic segmentation architectures.
We utilize both direct input, which adapts the input stem of feature extraction architectures to accept DMP channels, and hybrid architectures, a dual-stream design that fuses RGB and DMP encoders.
Using the iSAID benchmark dataset, we evaluate a variety of DMP differentials and structuring element shapes to more effectively provide shape information to the model.
Our results show that while non-DMP models generally outperform the direct-input variants, hybrid DMP consistently outperforms direct-input and is capable of surpassing a non-DMP model on mIoU, F1, and Recall.
\end{abstract}

\begin{IEEEkeywords}
Remote sensing, deep learning, semantic segmentation
\end{IEEEkeywords}

\section{Introduction}
\label{sec:intro}

\IEEEPARstart{S}{emantic} segmentation of remote sensing images is a critical technology underpinning numerous applications including environmental monitoring, urban planning, and rapid disaster response.
The ability to assign pixel-level semantic labels allows practitioners to map roads, identify damaged infrastructure, and monitor land use at a level of granularity that is not possible with coarser image classification or object detection alone.
However, the size and complexity of remote sensing imagery introduces significant challenges.
Datasets often include vast collections of high-resolution images, making manual analysis time consuming and labor intensive.
The sheer volume of data necessitates automated methods to extract information efficiently.
One approach to achieving this automation is the use of computer vision techniques, particularly via deep neural networks.
Convolutional neural networks (CNNs) and, more recently, transformer neural networks have achieved state of the art results across computer vision tasks in many domains, including semantic segmentation.
However, these networks are typically benchmarked on ground-level, natural images such as those found in the COCO~\cite{lin2014microsoft-MSCOCO} and ADE20K~\cite{zhou2017scene-ade20k} datasets.
These datasets do not fully capture the unique characteristics of overhead imagery such as large variations in object scales, severe imbalance between foreground and background pixels, and the presence of many small objects.
Consequently, models that excel on standard computer vision benchmarks may underperform on remote sensing tasks, leading to the development of specialized architectures.
Another emerging approach combines the powerful feature-learning abilities of neural networks with domain specific, hand-crafted features.
For remote sensing, the differential morphological profile (DMP)~\cite{pesaresi2001new_og_dmp} is one such feature.
The DMP is a method of shape extraction based on grayscale morphological operations and was originally designed for satellite image segmentation.
Recent work has explored the incorporation of the DMP into neural network architectures, which have been dubbed DMPNets.
DMPNets have seen success in remote sensing scene classification and object detection where the incorporation of shape information proved to be beneficial~\cite{scott2020differential_ogdmpnet, hurt2021improved, hurt2021differential_dmpfrcnn, hurt2022evolutionary, hurt2023hybrid}.
However, DMPNets have yet to be applied to remote sensing semantic segmentation.
This represents a notable research gap, as segmentation is the task for which the DMP was originally conceived and where its shape information could be most impactful to model performance.
In this work, we present the first study of DMPNets to remote sensing semantic segmentation.
Specifically, we investigate the incorporation of the DMP into three architectures: SegFormer~\cite{xie2021segformer}, SegNeXt~\cite{guo2022segnext}, and EfficientViT~\cite{cai2023efficientvit}.
We explore two strategies: (i) a \emph{Direct-Input} design, which alters each network's input layers to accept the channels produced by the DMP, and (ii) a \emph{Hybrid DMP} design which processes RGB and DMP inputs through separate encoders before fusing each learned representation.
Furthermore, we compare multiple sets of DMP differentials based on prior DMPNet studies.
We evaluate our proposed DMPNets for segmentation against the iSAID~\cite{waqas2019isaid} dataset, a large-scale remote sensing segmentation dataset covering $15$ object classes across varied geographical locations, environmental conditions, and sensor types.

The rest of this paper is organized as follows: 
In Section~\ref{sec:related_work}, we review related works to our proposed DMPNets for the segmentation task.
Then, in Section~\ref{sec:dmp_seg_nets}, we introduce our proposed network architectures and modifications in detail to enable standard segmentation networks to better integrate shape features.
Next, in Section~\ref{sec:experiments}, we evaluate our proposed DMPNets against the iSAID benchmark and analyze our results alongside state-of-the-art segmentation models, before presenting our conclusions and future work in Section~\ref{sec:conclusion}.

\section{Related Work}
\label{sec:related_work}
Image segmentation is a fundamental task in computer vision, with broad applications including medical image analysis, autonomous driving, and remote sensing.
Below, we review relevant families of segmentation methods, with emphasis on convolutional and transformer-based approaches, architectures designed specifically for remote sensing, and the integration of the DMP into neural networks.

\subsection{Convolution-Based Segmentation Networks}
\label{subsec:cnn_networks}
Following the success of the AlexNet~\cite{krizhevsky2012imagenetalexnet}, convolutional neural networks quickly became dominant in computer vision applications.
For semantic segmentation, the Fully Convolutional Network (FCN)~\cite{long2015FCN} was the first architecture to be trained as a convolution-only network end-to-end for the task of semantic segmentation.
Crucially, FCN leveraged encoders pretrained on the ImageNet dataset~\cite{deng2009imagenet}, replacing fully connected layers with convolution ones, taking advantage of their strong feature extraction capabilities.
Building upon FCN, the U-Net architecture~\cite{ronneberger2015u-unet} introduced an encoder-decoder design with skip connections, enabling the mingling of low-level and high-level features to produce precise segmentations.
Originally designed for medical image analysis, the U-Net is still a popular architecture, particularly for tasks with limited labeled data.
The DeepLab series~\cite{chen2014semantic-deeplabv1, chen2017deeplab-deeplabv2, chen2017rethinking-deeplabv3, chen2018encoder-deeplabv3+} further advanced CNN-based segmentation techniques by introducing architectural changes such as the use of dilated convolutions and conditional random fields for boundary refinement.
Like the FCN, DeepLab networks also took advantage of pretrained encoders such as Residual Networks~\cite{he2016deepresnets} and Xception~\cite{chollet2017xception}.
Although more recent architectures have expanded beyond pure CNNs, many contemporary networks retain core principles from these models such as encoder–decoder structures and multi-scale feature extraction.

\subsection{Transformer-Based Segmentation Networks}
\label{subsec:transformer_networks}
The transformer architecture~\cite{vaswani2017attention} revolutionized natural language processing by modeling long-range relationships with the self-attention operation.
The transformer was soon extended to computer vision with the Vision Transformer (ViT)~\cite{dosovitskiy2020image16x16vit}, which considered image patches as tokens as words were in natural language processing.
While other attempts had been made to port transformers to computer vision, the ViT was the first to demonstrate competitive performance with state of the art CNNs on large-scale image classification.
In semantic segmentation, transformers were first successfully adopted in architectures such as SETR~\cite{zheng2021rethinkingsetr}, which adopted a ViT encoder and applied a CNN-based decoder, and Segmenter~\cite{strudel2021segmenter} which also used a ViT encoder but applied another transformer in the decoder as well.
While transformer segmentation networks like SETR and Segmenter relied on the use of a ViT backbone and focused on the development of segmentation decoders, the SegFormer~\cite{xie2021segformer} architecture introduced a novel encoder and decoder.
Its Mix-Transformer (MiT) encoder produced hierarchical features of progressively lower resolutions unlike the ViT which maintains a single resolution, and adopted a more efficient self-attention operation.
Additionally, SegFormer's decoder only used lightweight fully connected layers.
By using these techniques, SegFormer achieved state of the art performance with fewer parameters than SETR and Segmenter.
While transformers achieved new state of the art results, the quadratic complexity of the self-attention mechanism with respect to the number of image patches remained a significant drawback.
The SegNeXt~\cite{guo2022segnext} architecture addressed this issue by utilizing convolution-based attention in its encoder, the Multi Scale Convolutional Attention Network (MSCAN).
By maintaining the overall transformer structure and replacing self-attention with convolutional attention, SegNeXt achieved new state of the art results with a significant reduction in computational cost compared to previous transformer designs.
The EfficientViT~\cite{cai2023efficientvit} architecture, took a different approach to addressing self-attention's complexity.
The softmax-based similarity function in self-attention was replaced with one based on the ReLU~\cite{nair2010rectified-relu} activation function.
As a result, this altered self-attention operation became quadratic with respect to the embedding dimension, $O(nd^{2})$, rather than the number of tokens, $O(n^2d)$.
This resulted in a cheaper operation, as typically the number of patches is much greater than the embedding dimension.
EfficientViT produced state of the art results while achieving inference speeds much faster than previous models, especially on mobile devices.

\subsection{Semantic Segmentation Architectures for Remote Sensing}
\label{subsec:rs_seg_networks}
While the CNN and transformer-based segmentation models described in Sections~\ref{subsec:cnn_networks} and~\ref{subsec:transformer_networks} have shown strong results on natural image benchmarks, remote sensing imagery presents unique challenges.
These include extreme variation in object size, high intra-class variability, severe foreground–background imbalance, and many small objects. These differences have motivated specialized architectures for remote sensing segmentation.
The ResUNet-a architecture~\cite{diakogiannis2020resunet} combined the strengths of residual networks and the U-Net while utilizing multi-task training to increase performance on remote sensing imagery.
In its encoder, residual connections and dilated convolutions are used to avoid vanishing gradients and improve multi-scale feature extraction.
When producing predictions, the model follows a sequential, multi-task approach: first object boundaries are predicted, then the Euclidian distance transform, then the actual segmentation mask, and finally a color reconstruction of the input image.
Additionally, a new loss function is introduced, Tanimoto loss, which aids in convergence when training on data with severe class imbalance.
To address poor performance of segmentation networks on small objects images, the Foreground Activation-Driven Small Object Semantic Segmentation (FactSeg)~\cite{ma2021factseg} network was proposed.
FactSeg utilizes a dual-branch encoder design with one branch being dedicated to distinguishing foreground from background and the other for refining small details in the objects identified by the foreground-background branch.
Its training also employs a modification of Online Hard Example Mining~\cite{shrivastava2016training-ohem} technique which focuses on difficult, small objects.
The Foreground-Aware Relation Network (FarSeg / FarSeg++)~\cite{zheng2020foreground-farseg, zheng2023farseg++} also sought to address large foreground-background imbalance by learning to model foreground pixels versus background pixels.
It features a multi-branch encoder based on the Feature Pyramid Network~\cite{lin2017feature-fpn} design which produces multi-scale image features and also learns a weighting for each scale, based on modeling the foreground objects and the background scene.
Additionally, a dynamic loss weighting is learned to mitigate the issue of class imbalance, similar to a Focal Loss~\cite{lin2017focal}.
More recently, the AerialFormer~\cite{hanyu2024aerialformer} architecture introduced transformer-based designs specifically tuned for remote sensing imagery.
It combines the strengths of CNNs and transformers by using Swin~\cite{liu2021swin} encoder blocks as well as a CNN-stem to capture finer, local details at a higher resolution to complement the global features produced by the transformer features and aid in segmenting small objects.
Overall, AerialFormer also follows a U-Net-like structure, allowing for earlier contexts from the encoder to persist into the decoder.

\subsection{Differential Morphological Profile}
\label{subsec:dmp}
The differential morphological profile (DMP) is a method of extracting multi-scale shape information from an image using grayscale morphology, proposed by Pesaresi and Benediktsson~\cite{pesaresi2001new_og_dmp}.
In the original formulation, the DMP is constructed using geodesic structuring elements, but here we adopt a simplified version with flat structuring elements to reduce computational demand.
In grayscale morphology, the two base operations are dilation ($\delta$) and erosion ($\epsilon$).
They are defined as:
\begin{align}
    \delta^{SE}(I) = \delta(I) \vee \mathtt{SE}
    \\
    \epsilon^{SE}(I) = \epsilon(I) \wedge \mathtt{SE}
\end{align}
where $I$ is a single-banded image, $SE$ is the structuring element, $\vee$ is a set-wise maximum, and $\wedge$ is a set-wise minimum.
Using these two base operations, we can build the opening $\gamma$ and closing $\varphi$ operations:
\begin{align}
    \gamma^{SE}(I) = \delta^{SE}(\epsilon^{SE}(I))
    \\
    \varphi^{SE}(I) = \epsilon^{SE}(\delta^{SE}(I))
\end{align}
The morphological profile of an image can then be computed by performing openings and closings at strictly increases structuring element sizes:
\begin{align}
    \Pi\gamma(I) = \{\Pi\gamma_l : \gamma^{SE=l}(I), \forall l \in L\}
    \\
    \Pi\varphi(I) = \{\Pi\varphi_l : \varphi^{SE=l}(I), \forall l \in L\}
\end{align}
Finally, differential morphological profile comprises the absolute value of the differences between opening and closing profiles at the increasing structuring element sizes:
\begin{align}
    \Delta\gamma(I) = \{\Delta\gamma_l : \Delta\gamma_l = |\Pi\gamma^{SE=l}(I) - \Pi\gamma^{SE=l-1}|, \forall l \in L^{\prime}\}
    \\
    \Delta\varphi(I) = \{\Delta\varphi_l : \Delta\varphi_l = |\Pi\varphi^{SE=l}(I) - \Pi\varphi^{SE=l-1}|, \forall l \in L^{\prime}\}
\end{align}
where $L^{\prime}$ = $L \setminus min(L)$.
The DMP has been widely used in remote sensing as a hand-crafted feature extractor, particularly for segmentation and classification tasks where geometric context is critical.
Its multi-scale nature makes it well suited for handling the diverse object sizes common in aerial imagery.

\subsection{Differential Morphological Profile Neural Networks}
\label{subsec:dmp_nets}
\subsubsection{DMPNet}
The original DMPNet~\cite{scott2020differential_ogdmpnet} from Scott et al. utilized shape extraction features from the DMP as a direct input to a classification network.
An input image is first reduced to a single band with a learnable point-wise convolution, making it suitable for grayscale morphological operations.
The opening and closing profiles for the grayscale image are calculated for some predefined set of structuring element sizes, and their differentials are computed, resulting in the DMP.
Notably, a flat structuring element is used for computational efficiency compared to the geodesic structuring element utilized in the original conception of the DMP.
The DMP opening channels and closing channels are each reduced to a single channel with their own point-wise convolution.
Finally, the opening channel, the grayscale image, and the closing channel are concatenated, resulting in a 3-channel tensor.
The 3-channel tensor is then fed into an encoder, which was chosen to be a VGG16~\cite{simonyan2014very_vggnets} network for DMPNet.
For the DMP shape extraction, a square structuring element was chosen, with $\{3,5,7,9\}$ as the sizes.
Note that the curly-brace notation implies differentials of $[5-3], [7-5], [9-7]$.
These two forms of noting the selected DMP differentials are used interchangeably in this work.
Experiments were performed to compare DMPNet's performance on remote sensing scene classification to a VGG16 without any DMP integration.
The Benchmark MetaDataset V2~\cite{hurt2020enabling_mdsv2dataset} (MDSv2) was used for training and evaluation, which contains approximately $85,000$ images across $33$ classes.
For both the standalone VGG16 model and the VGG16 model used as DMPNet's encoder, the model was initialized with ImageNet pre-trained weights.
DMPNet produced F1-scores competitive with VGG16 for most classes on the MDSv2 test set, demonstrating that a network with DMP integration can learn useful features for remote sensing scene classification.
Specifically, the average F1-score of DMPNet only lagged behind VGG16 by approximately $1$ percentage point.
The authors found that DMPNet outperformed VGG16 on the \textit{Crosswalk} and \textit{Oil Well} classes and attributed this to the small structures present such as the painted crosswalk lines and the shadows cast by the oil well pipes.
Given the smaller set of structuring element sizes, these patterns were successfully extracted by the DMP and aided in DMPNet's ability to classify them.
In contrast, VGG16 outperformed DMPNet most significantly for classes where large structures are key to identifying the target category such as \textit{Church} and \textit{Ferry Terminal}.

\subsubsection{Improved DMPNet}
A follow-up work from Hurt et al.~\cite{hurt2021improved}, made changes to the DMPNet architecture and used a different approach for experiments.
In an initial set of experiments with the MDSv2 dataset, the authors discovered that when no longer using ImageNet pre-trained weights, the gap between DMPNet and VGG16's F1-scores on the test set grew, with DMPNet now achieving an average F1-score approximately $7$ percentage points lower than VGG16, concluding that the use of ImageNet weights inflated the performance of the original DMPNet.
To more accurately demonstrate the contribution of the DMP features, the rest of the experiments laid out in the paper utilize encoders with randomly initialized weights.
Other than VGG16, a ResNet18, and MobileNetV2~\cite{sandler2018mobilenetv2} network were evaluated for the DMPNet encoder.
One architectural change was the use of depth-extended DMP features, i.e., removing the point-wise convolution to compress the opening and closing DMP channels to single bands, and instead concatenating the full set of opening and closing DMP channels with the grayscale input.
For the structuring element sizes of $\{3,5,7,9\}$, this produces a 7-channel input tensor to the encoder, requiring slight modification of the input layer for the encoders.
A 15-channel depth-extended variant is evaluated as well, with structuring element sizes $\{3, 5, 7, 9, 15, 21, 27, 35\}$.
Another change of note is that the point-wise convolution used to reduce the RGB input image to grayscale was replaced by an ITU-R 601-2 LUMA transform.
For the VGG16 DMPNet, the use of the 7-channel depth extended architecture closes the gap in average F1-score between DMPNet and VGG16 from $7$ percentage points to $0.14$, suggesting that the depth-extended architecture is a vital to successful use of DMP features in a neural network.
Improved performance from using a 7-channel depth-extended DMPNet held for the model with a MobileNetV2 encoder, but not for the model with a ResNet18 encoder.
However, when using the 15-channel depth-extended architecture, the ResNet18 DMPNet achieved a higher F1-score than the non-depth extended 3-channel variant and also achieved a higher F1-score than the standalone ResNet18 model and was the only DMPNet to do so.
The varying results suggested that per-model tuning is likely a requirement for DMPNet, i.e., there is no one-size-fits-all set of structuring element sizes or number of input channels which will yield improved performance for every model.

\subsubsection{DMP-FRCNN}
In another paper centered on maneuverability hazard detection in overhead unmanned aerial system (UAS) imagery~\cite{hurt2021differential_dmpfrcnn}, DMPNet was used as the backbone of a Faster R-CNN architecture~\cite{ren2015fasterrcnn}.
The dataset used for training and evaluation contained over $1,300$ images captured from a UAS at a 60 meter altitude, providing high quality captures of a training area.
The annotations were at the object bounding box level and more than $7,200$ bounding boxes were labeled.
Sixteen classes were present in the dataset related to maneuverability hazard detection such as cone, rubble, civilian vehicle, military vehicle, jersey barrier, etc.
The UAS images were captured along four routes within the training area, going down and back, and three were chosen for training and one was chosen for the holdout test set.
Like in the improved DMPNet publication, a depth-extended DMPNet architecture was used and the VGG16, ResNet18, and MobileNetV2 architectures were evaluated for the DMPNet encoder.
A VGG16 architecture with batch normalization incorporated was also used and is referred to as VGG16-BN.
The \se size pairs of $\{[9,3], [5,3], [9,5], [13,5]\}$ were manually selected based on the training data.
The DMP features are passed through the encoder, whose features are then used as input to Faster-RCNN's region proposal network.
The Faster-RCNN components from the region proposal network and onward are the same as the original architecture other than the use of the Region-Of-Interest-Align from the Mask-RCNN paper~\cite{he2017maskrcnn} instead of Region-Of-Interest Pool from the Faster-RCNN paper.
Unlike earlier DMPNet classification experiments, there was mostly a significant drop in performance when switching to a DMPNet backbone.
For VGG16-BN, ResNet18, and MobileNetV2, DMPNet had a difference of validation mAP50 of $-10.01\%$, $-9.65\%$, and $-13.97\%$, respectively.
However, for the VGG16 encoder without batch normalization, the DMPNet backbone produced a validation mAP50 difference of $+0.96\%$, suggesting that batch normalization has a significant effect on the use of DMP features.
Despite the worse average performance on mAP50, of the $16$ classes present, $7$ of them had a best per-class AR50 from a DMPNet backbone, implying that DMP features may still be useful for certain classes.

\subsubsection{Evolutionary Search for DMP Parameters}
Up to this point, structuring element (SE) sizes for the DMP were manually selected by researchers, a process that is subjective and unlikely to scale to large datasets.
To address this, Hurt et al.~\cite{hurt2022evolutionary} applied an evolutionary computation algorithm (ECA) to automatically optimize SE differentials for DMP-based neural networks.
The study evaluated whether such an approach could discover parameter sets that improve object detection performance on overhead imagery of airplanes.

In this formulation, each chromosome encoded a set of DMP differentials, and the fitness of a chromosome was defined by the F1-score of a DMP-FRCNN detector trained on the RarePlanes dataset~\cite{shermeyer2021rareplanes}.
The population evolved over 20 generations through selection, crossover, and mutation, with crossover and mutation probabilities set to 0.95 and 0.02, respectively.
Candidate SE sizes were restricted to odd values between 3 and 31, consistent with the range of aircraft sizes present in the dataset.
Results showed that the best fitness score improved from 878.87 in the initial generation to 883.02 in the final generation, indicating that the ECA was able to discover non-trivial SE configurations that enhanced DMP-FRCNN performance.
Analysis revealed that the best chromosome fitness steadily improved, while worst-case and average fitness values fluctuated more noisily across generations.
This suggests that evolutionary search can identify effective parameterizations that may not be obvious through manual design.

\subsubsection{Hybrid DMPNet}
In the prior DMP publications, the features that were sent into the network were the bands produced by DMP shape extraction and the original input image after being cast to grayscale. 
However, this method had the tradeoff of losing spectral information, i.e., color information was not considered.
To attempt to remedy this tradeoff, a Hybrid DMP approach was proposed by Hurt et al~\cite{hurt2023hybrid}.
In Hybrid DMP, there are two independent convolutional feature extractors.
One encoder processes the RGB input image as in non-DMP feature extraction.
The other encoder receives the DMP shape extraction bands and grayscale-cast input image as in prior DMPNet architectures.
After the features have been independently produced from each encoder, a feature fusion method is applied to combine the features from each.
In the published work, a concatenation along the channel dimension was used.
The output of the feature fusion is used as the input to a task-specific head for tasks such like classification or object detection.
In the publication, the features were used as input to a region proposal network and Faster RCNN head.
The Hybrid DMP FRCNN model was evaluated on the RarePlanes and DOTA datasets.
DOTA~\cite{xia2018dota} is another object detection dataset in the domain of overhead images, however it has 16 classes in a variety of categories such as \textit{plane}, \textit{ship}, \textit{tennis court}, etc.
The AP50, AR50, and Weighted Optimal F1 Score ($WF_{1}^{Opt}$) were used as metrics for evaluating models.
AP50 and AR50 were calculated using the \textit{pycocotools}~\cite{pycocotools} library.
The $F_{1}^{Opt}$ is produced by calculating the $F1$ score across predictions at different bounding box confidence thresholds, and picking the confidence threshold which yields the highest $F1$ score.
Then, the $WF_{1}^{Opt}$ can be computed by taking the $F1$ score at the optimal threshold for each class, weighting each $F1$ score by the number of samples of that class, and then averaging across the classes.
VGG16, ResNet18, and MobileNetV2 were selected as the encoders for the Hybrid DMP FRCNN experiments.
These older, more shallow networks were chosen to highlight how DMP features can aid in bolstering the performance of networks which are simpler and potentially have less capacity to learn high level features.
For each encoder, four experiments are performed.
The first is a single-branch FRCNN where the RGB input image is simply passed through the encoder and then onto the FRCNN-specific layers.
The second experiment is the same as the first, but using the RGB image cast to grayscale as input.
The third experiment follows the original DMP-FRCNN architecture, i.e., the input to the network is the DMP shape extraction bands as well as the input image cast to grayscale.
The fourth experiment is the Hybrid DMP FRCNN as described above.
In the RarePlanes experiments, it was seen that all four models achieved relatively similar metrics, demonstrating that DMPNet and Hybrid DMP can achieve object detection results which are competitive with a vanilla RGB- or gray-input network.
However, in general, the DMP-enabled models lagged behind non-DMP models except for in a few instances.
Conversely, the DMP-enabled models showed better performance in the DOTA experiments, particularly for the Hybrid DMP models.
For example, Hybrid DMP FRCNN achieved the best AR50 in each of the 4-experiment groups for the 4 encoders.
Additionally, the Hybrid DMP with a ResNet18 encoder achieved the best AP50, AR50, and $WF_{1}^{Opt}$ in the ResNet18 experiment group.
This result suggested that the Hybrid DMP approach can successfully utilize both DMP and RGB features to achieve better performance than relying on one type of feature alone.

\section{Differential Morphological Profile Semantic Segmentation Networks}
\label{sec:dmp_seg_nets}
While DMPNet method has been applied to the classification and object detection tasks, there exists a gap in research applying DMPNets to the task of semantic segmentation. 
We hypothesize that the DMP is an ideal candidate for enhancing segmentation networks due to its shape extraction ability and its original use as a method of satellite image segmentation.
In this section, we describe how DMP segmentation networks will be created.

\subsection{Selection of Structuring Elements}
\label{subsec:se_selection}
Successful shape extraction via the DMP is highly sensitive to selection of structuring elements (\secompact).
The size of the structuring elements is crucial as is the size of the differentials.
Here we outline the structuring elements selected for experiments, based on structuring element selections in prior DMPNet research.

\subsubsection{Original DMPNet Differentials}
\begin{figure*}
    \centering
    \subfloat[Input Image]{
        \includegraphics[width=0.22\textwidth]{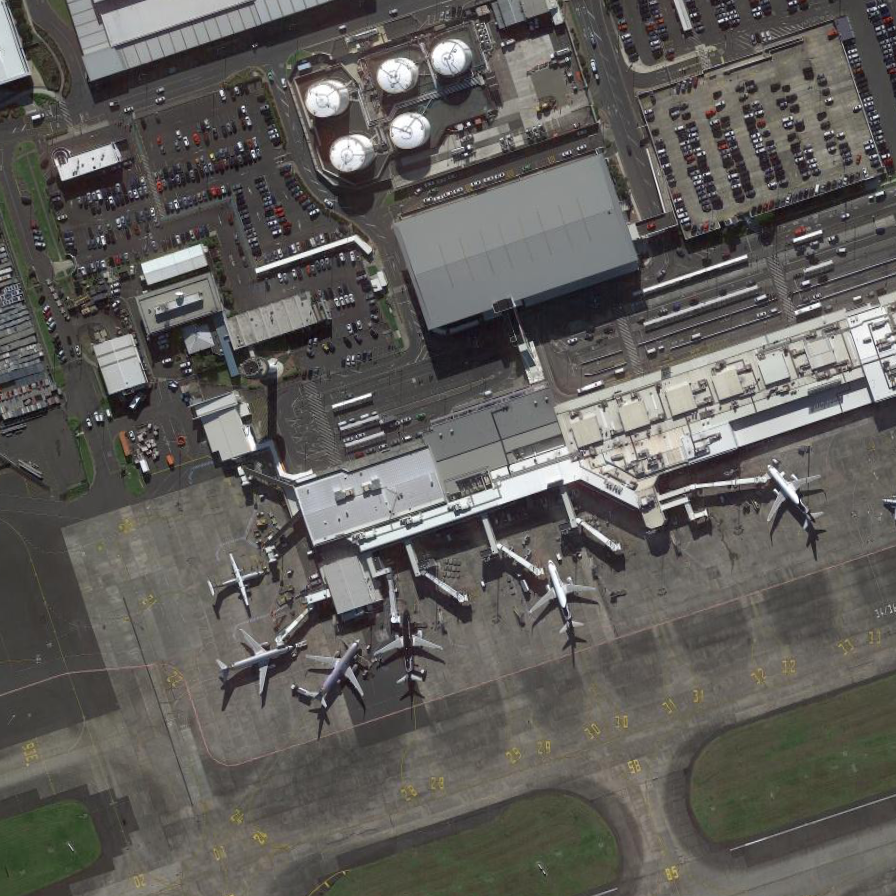}
    } \hfill
    \subfloat[$5-3$ Closing]{
        \includegraphics[width=0.22\textwidth]{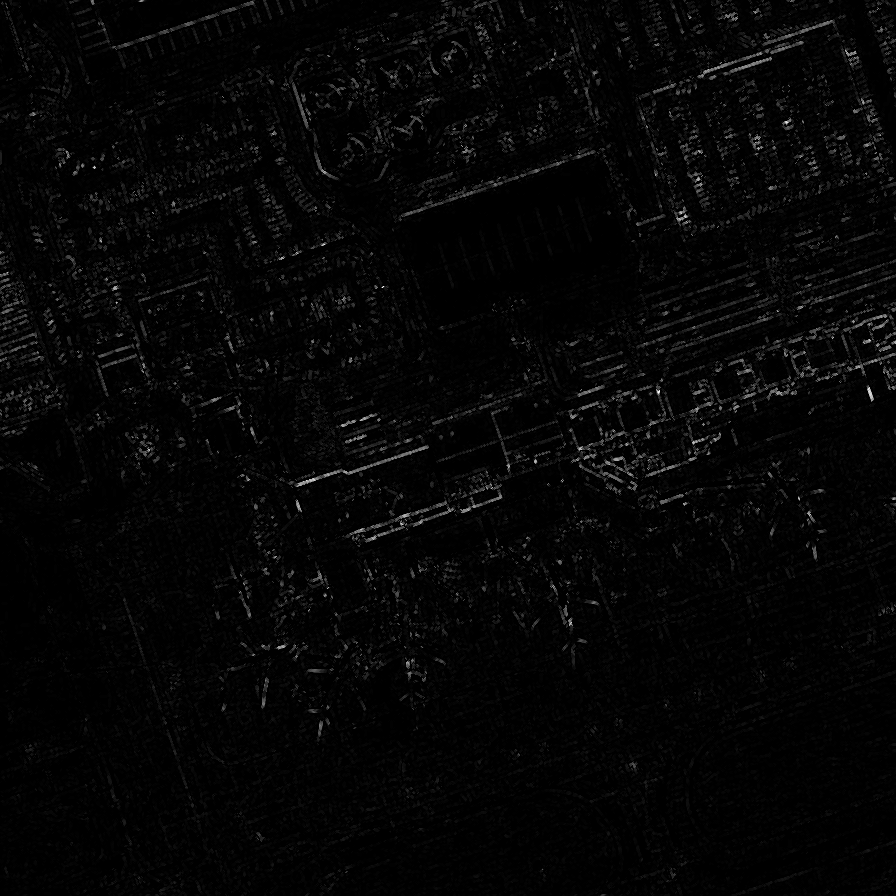}
    } \hfill
    \subfloat[$7-5$ Closing]{
        \includegraphics[width=0.22\textwidth]{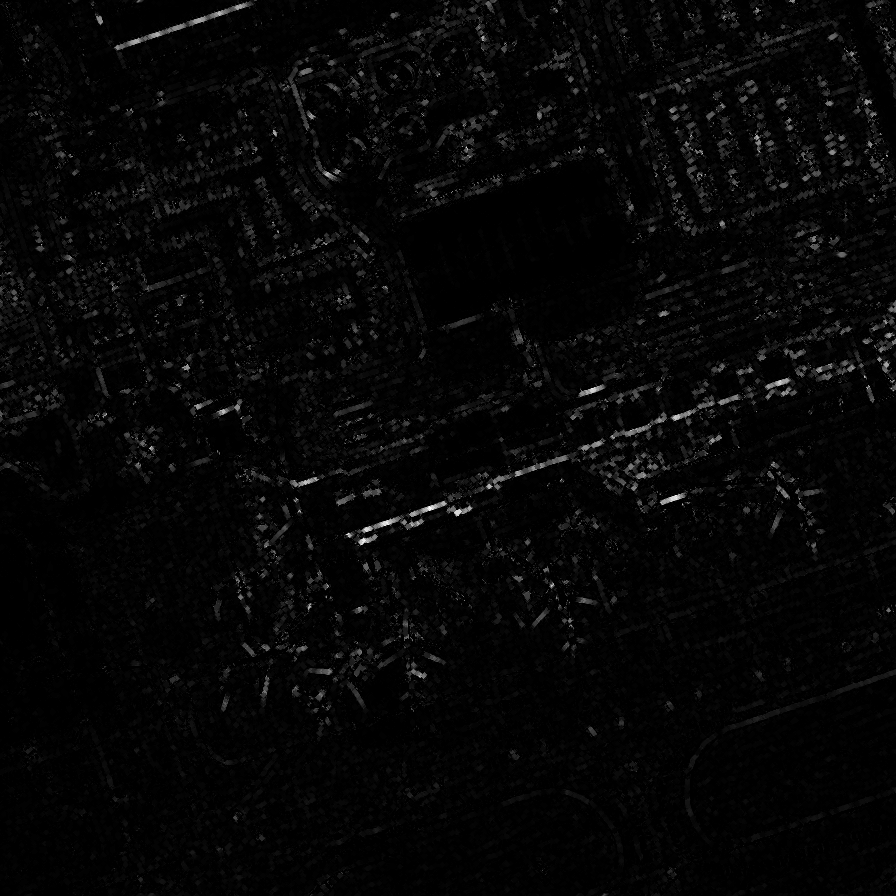}
    } \hfill
    \subfloat[$9-7$ Closing]{
        \includegraphics[width=0.22\textwidth]{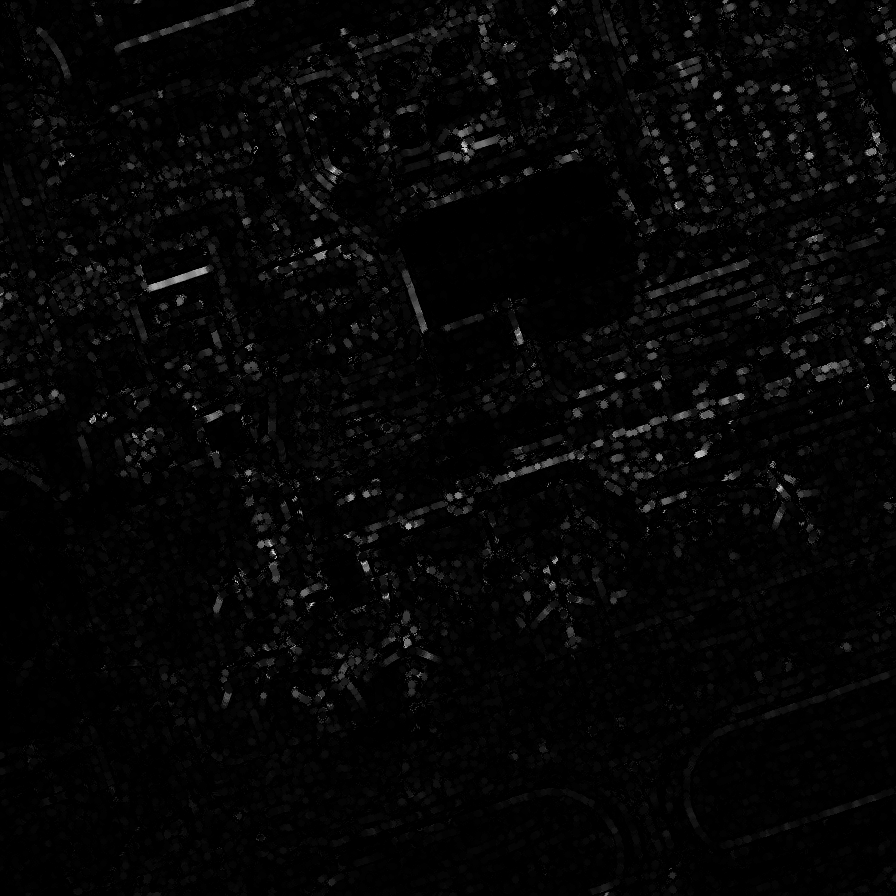}
    }
    \caption{Closing profile using the $[5-3], [7-5], [9-7]$ differentials.}
    \label{example_small_closing_profile}
\end{figure*}

\begin{figure*}
    \centering
    \subfloat[Input Image]{
        \includegraphics[width=0.22\textwidth]{overhead_ex1.png}
    } \hfill
    \subfloat[$5-3$ Opening]{
        \includegraphics[width=0.22\textwidth]{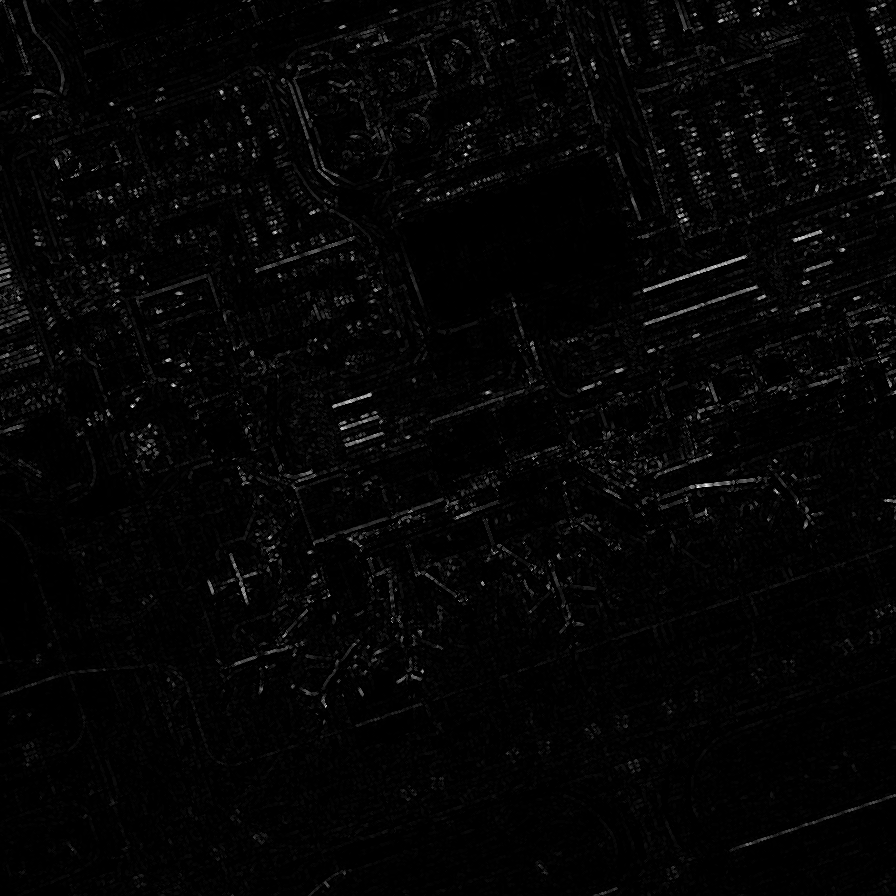}
    } \hfill
    \subfloat[$7-5$ Opening]{
        \includegraphics[width=0.22\textwidth]{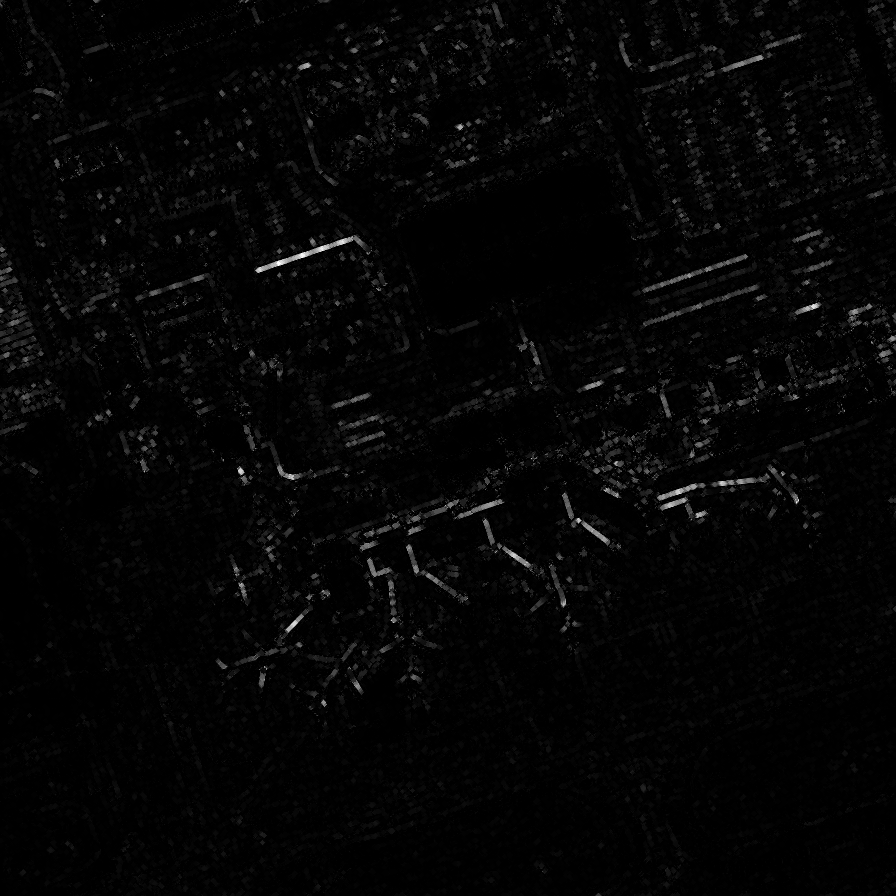}
    } \hfill
    \subfloat[$9-7$ Opening]{
        \includegraphics[width=0.22\textwidth]{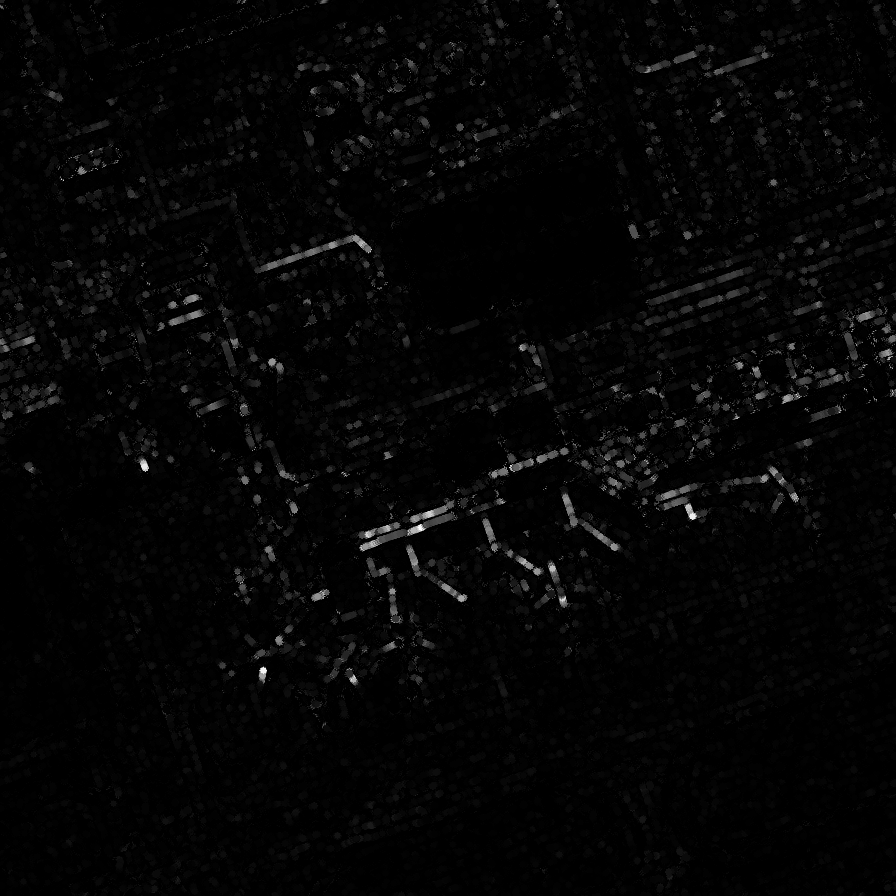}
    }
    \caption{Opening profile using the $[5-3], [7-5], [9-7]$ differentials.}
    \label{example_small_opening_profile}
\end{figure*}

In the original DMPNet~\cite{scott2020differential_ogdmpnet}, the \se differentials of $$[5-3], [7-5], [9-7]$$ were utilized.
The use of relatively small \se sizes results in a DMP which is computationally efficient.
Additionally, the anticipated result of using these sizes is that the DMP would highlight small, thin structures within an image.
For instance, the \textit{crosswalk} class saw improved performance in the aforementioned publication.
Crosswalks are often bright, thin, parallel lines which would have been more successfully extracted by the DMP with these \se sizes.
In the context of semantic segmentation, this set of sizes could be beneficial for highlighting fine-grained details such as object boundaries and small objects.
The primary limitation of this set of sizes is its restriction to extracting details at smaller scales.
While small details are important, they are only one part of the picture, and for scenes which are dominated by large structures, this set of \se sizes might only emphasize texture or minor internal details, failing to provide a holistic representation of the objects themselves.
Therefore, this could hinder a neural network's ability to understand the shape of large objects, potentially leading to many missed pixel detections.
Figures~\ref{example_small_closing_profile} and~\ref{example_small_opening_profile} demonstrate the opening and closing profiles using the $[5-3], [7-5], [9-7]$ differentials.
In the closing profile, darker regions such as object shadows and darkly colored vehicles show higher responses.
In the opening profile, lighter regions such as jet bridges, aircraft, and storage tanks show higher responses.
%

\subsubsection{Improved DMPNet Differentials}
In the Improved DMPNet publication~\cite{hurt2021improved}, the \se differentials used were $$[5-3], [7-5], [9-7], [15-9], [21-15], [27-21], [35-27]$$ producing a 15-band input.
This expanded set of \se sizes presents a significantly broader set of scales for the DMP compared to the set used in the original DMPNet.
The multi-scale shape context which is extracted by the DMP with this set of sizes is theoretically more aligned with the task of semantic segmentation, maintaining the small \se sizes for small details, while introducing larger sizes for the extraction of whole objects.
In the publication, the use of this set of sizes had varying results across different encoders, however the DMPNet using these sizes and a ResNet18~\cite{he2016deepresnets} encoder was the first DMPNet to surpass the RGB-only baseline in classification F1-score.
The primary drawback of this approach is the increased computational cost associated with the use of larger \se sizes.
This increases processing time which may not be suitable for certain applications such as disaster response.
Additionally, with more DMP differentials overall, there is the risk of introducing redundant information if some of the scales do not provide distinct information for a specific image.

\subsubsection{Evolutionary Computation-Discovered Differentials}
In the work using evolutionary computation algorithms (ECA)~\cite{hurt2022evolutionary}, two sets of optimal DMP parameters were produced which improved object detection performance on the RarePlanes dataset~\cite{shermeyer2021rareplanes}.
The sets were:
$$[29-5], [23-5], [19-13], [17-13], [17-9], [15-11], [13-7]$$
and
$$[29-5], [23-9], [23-5], [19-13], [17-13], [15-11], [13-7]$$
A benefit of these ECA-discovered \se sets is their task-specific optimization.
Through the continuous evolution of DMP differentials over several generations, \se sizes are selected which are more likely to capture useful features that are directly beneficial to the target task.
In the two sets, the lack of differentials where both \se sizes are small is an interesting result and somewhat non-intuitive.
This suggests that ECAs can uncover more complex relationships between objects, \se sizes, and their usefulness to a neural network that may not be apparent when performing manual selection.
However, a significant weakness of the use of ECAs for \se selection is the risk of overfitting to the specific characteristics of the task and dataset which were used for optimization.
It is not guaranteed that these sets of \se sizes will generalize to both the distinct task of semantic segmentation and to a new dataset.
Additionally, the black-box nature of the DMPNets which will use the ECA-discovered sizes makes it difficult to reason about why a particular set of sizes is useful.
Furthermore, the cost of \se size search via an ECA can be very high.
For example, in the ECA DMPNet publication, $32$ days of training were required to complete the ECA process, although this process is only required once.
Finally, as in the case of the Improved DMPNet, the use of larger \se sizes also leads to increased computational cost during the generation of the DMP compared to the sizes used in the original DMPNet.

\subsection{Methods of DMP Incorporation}
Beyond the selection of \se sizes, the method by which features produced by the DMP are integrated into a neural network is another critical design consideration.
The selected strategy dictates how and at what stage the DMP information interacts with other information such as color or if that interaction occurs at all.
Different strategies introduce varying levels of complexity, computational cost, and the types of features which a neural network is given to learn from.
This section reviews the incorporation methods considered in this research.

\label{subsec:dmp_incorporation}
\subsubsection{Direct-Input}
\begin{figure*}[tbp]
    \begin{center}
        \includegraphics[width=1\textwidth]{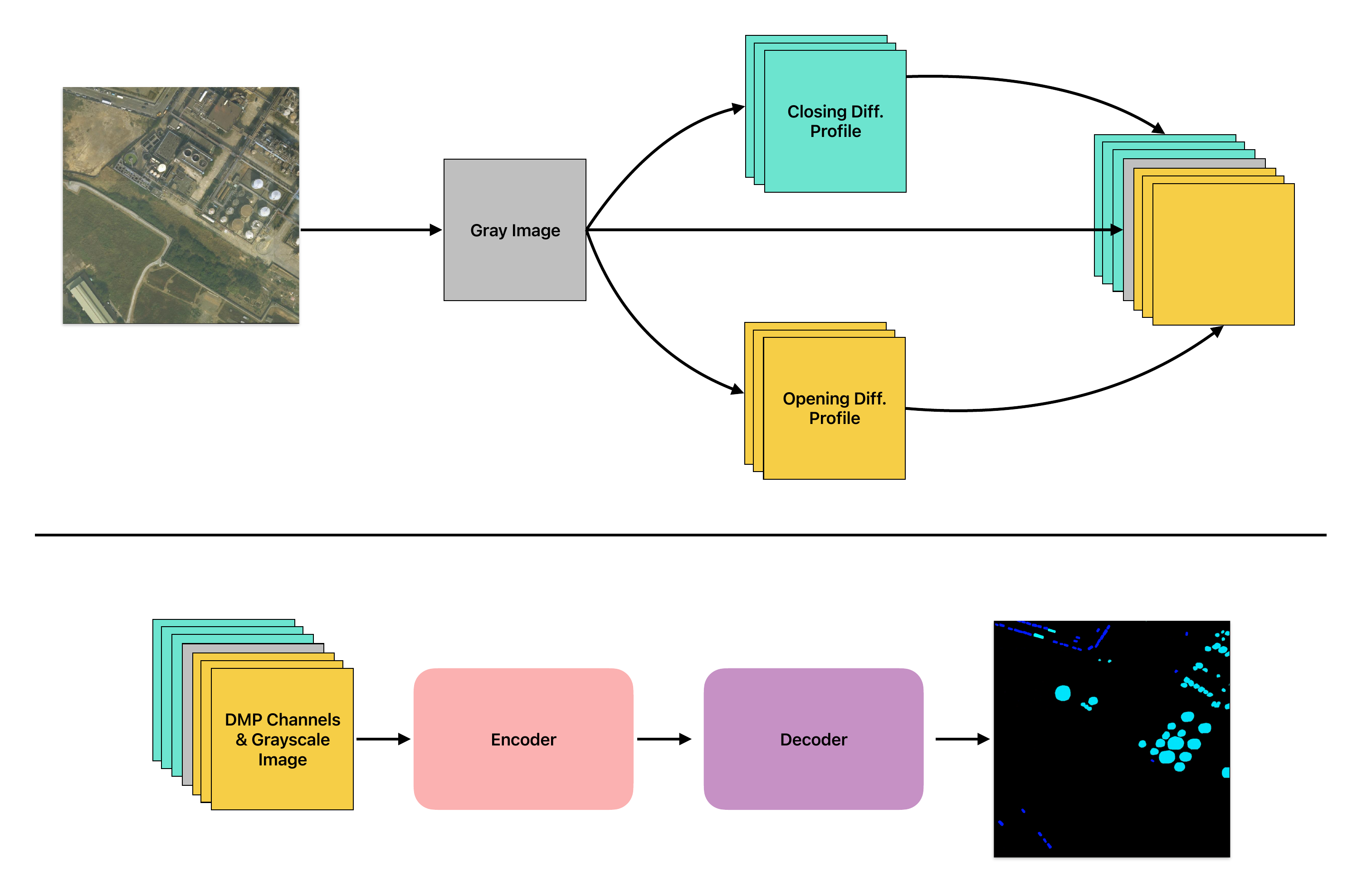}
    \end{center}
\caption{The Direct-In DMP incorporation method.}
\label{fig:direct_in_arch}
\end{figure*}

Following previous DMPNet research~\cite{scott2020differential_ogdmpnet, hurt2021improved, hurt2021differential_dmpfrcnn, hurt2023hybrid}, a direct input method of DMP incorporation is considered.
This approach involves preprocessing the input image to generate DMP features and then concatenating them with the image converted to grayscale to form the input tensor for the encoder.
In particular, the depth-extended design is adopted from the Improved DMPNet publication:
\begin{enumerate}
\item Image received as input.
\item Image cast to grayscale.
\item DMP generated from the grayscale image (both opening and closing profiles).
\item The closing profile, the original grayscale image, and the opening profile are concatenated along the channel dimension. For instance, if using 7 \se differentials, this results in a (7 + 1 + 7) = 15-band tensor.
\item This composite tensor is then fed as input to the encoder.
\item Encoder-produced features are input to the specific task head (in this case, a segmentation decoder).
\item Task head produces final predictions.
\end{enumerate}
See Figure~\ref{fig:direct_in_arch} for a visual representation of the direct-in method.
%

%
This method has the advantage of being relatively simple to implement, only requiring modification of the encoder at the input layer to expect the appropriate number of input channels.
By providing the DMP as input, the neural network can directly learn from the extracted multi-scale shape information.
The inclusion of the grayscale band ensures that information about pixel intensity, texture, and contrast are preserved and are processed alongside the DMP's structural information.
We hypothesize that for semantic segmentation, which is fundamentally about defining the exact shape of objects, this explicit shape information could be beneficial.
However, a significant weakness of this method is the loss of color information, as the RGB image is converted to grayscale before DMP generation and only the grayscale image is provided as additional input.
Color can provide powerful discriminating information for many classes, for example differentiating between mountains and ocean where there may be textural and structural similarities, but distinct color appearance.
As a result, only providing the DMP and grayscale bands may cause a neural network to struggle to distinguish between classes which are most distinct in their spectral appearance.
Another consideration is that by altering the initial input layer of the encoder to accept a new number of bands, the weights of the layer will be reset and randomly initialized or carefully adapted to introduce additional weights.
This could potentially reduce the benefits from pre-training for that specific layer which could then cascade throughout the network.
Increasing the number of input channels to the network also increases the computational cost, especially as the spatial resolution at the input layer of the network is largest, meaning more locations for convolutional filters to visit.
In the original DMPNet, a $1 \times 1$ convolution was used to compress the closing and profile bands to a single band each.
Stacked with the grayscale image, this maintained the standard 3-channel input and allowed the use of unaltered pre-trained encoder input layers.
However, this compression risks losing valuable information from the diverse DMP bands and previous research has shown this approach to yield inferior results to the depth-extended approach~\cite{hurt2021improved}, and was therefore not considered for this work.

\subsubsection{Hybrid DMP}

\begin{figure*}[tbp]
    \begin{center}
        \includegraphics[width=1\textwidth]{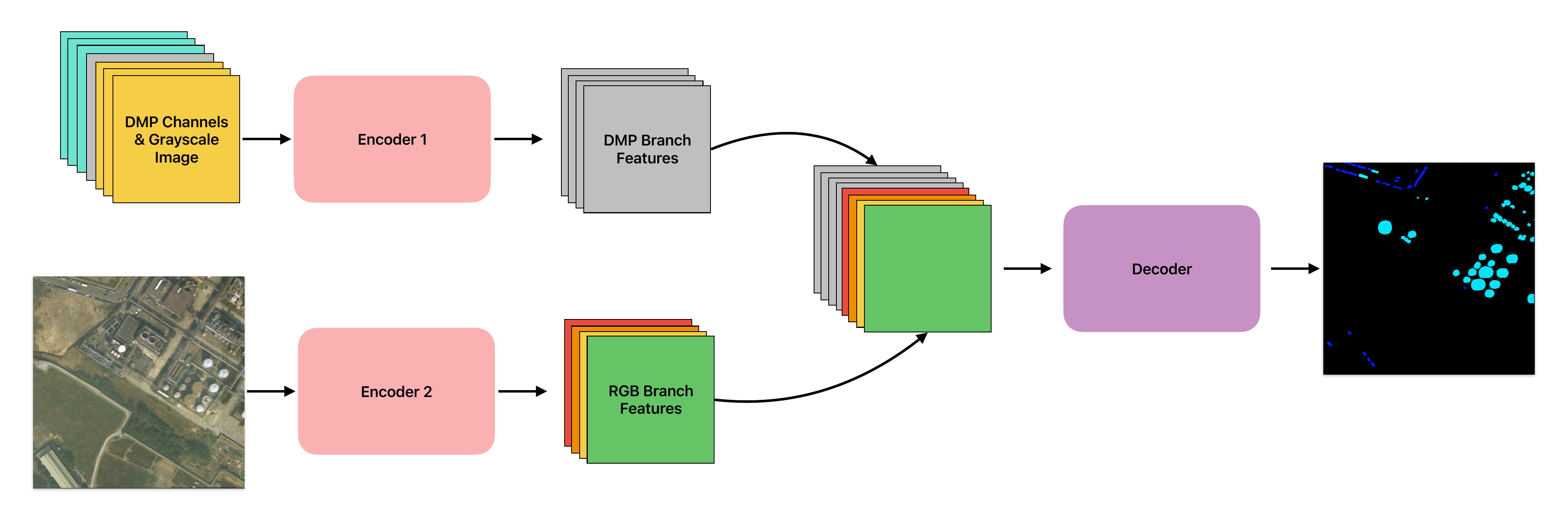}
    \end{center}
\caption{The Hybrid DMP incorporation method.}
\label{fig:hybridmp_arch}
\end{figure*}

While shape and texture information may aid in producing accurate segmentation maps, the loss of color information via the direct input method of DMP incorporation may limit performance.
The use of the Hybrid DMP~\cite{hurt2023hybrid} method of DMP incorporation is one approach to reintroducing color information back into the learning process.
By using two parallel branches, one for RGB features and one for DMP features, the model can learn to utilize information from both modalities when generating segmentation predictions.
In this work, the implementation from the publication introducing the Hybrid DMP is followed:
\begin{enumerate}
\item Image received as input.
\item RGB Stream: The original RGB input image is processed by the encoder.
\item DMP Stream: The input image is cast to grayscale and is used to generate the DMP, which is then concatenated with the grayscale image itself and processed by a separate encoder.
\item Feature Fusion: The feature maps produced by RGB and DMP streams are then concatenated along the channel dimension.
\item These fused features are then passed to the task-specific head (e.g., segmentation decoder).
\end{enumerate}
Note that the encoder architecture used in each stream is the same, although the weights are not shared between the streams.
Figure~\ref{fig:hybridmp_arch} displays the Hybrid DMP method.
%

%
By incorporating information from both modalities, the network can utilize the useful information provided by each.
The RGB branch may focus on spectral information while the DMP branch may specialize in learning from the shape information.
The fusion step allows for interaction between the feature sets, potentially leading to a richer representation of the objects in the input image to be used by the decoder.
Hybrid DMP does, however, introduce a significant computational cost.
By using dual encoders, the number of parameters and operations in the encoder network is approximately doubled.
As a result, training and inference time will increase as well as the GPU memory required for both.
Additionally, there are more hyper-parameters introduced which need to be considered such as if the encoders should be identical as well as the selection of fusion strategy.

\section{Experiments}
\label{sec:experiments}
Various experiments are run to evaluate the efficacy of the incorporation of the differential morphological profile into semantic segmentation networks.
Three network architectures are evaluated, SegFormer-B1~\cite{xie2021segformer}, SegNeXt-S~\cite{guo2022segnext}, and EfficientViT-B2~\cite{cai2023efficientvit}.
These particular networks were chosen because they offer a decent tradeoff between segmentation accuracy and computational cost.
Additionally, they were selected to have a similar number of parameters to the networks utilized in prior DMPNet research.
SegFormer-B1 has $13.7$M parameters, SegNeXt-S has $13.9$M, and EfficientViT-B2 has $15.3$M.
For each network, the DMP incorporation strategy and parameters from DMPNet, Improved DMPNet, and the evolutionary computation algorithm DMP-FRCNN are utilized.

\subsection{Dataset}
\label{subsec:isad_dataset}
For evaluation of each model and configuration, the iSAID\cite{waqas2019isaid} benchmark dataset is used.
It provides pixel-level annotations for $2,806$ overhead images across $15$ object categories.
The original format of the annotations are instance-level masks, however the authors provide a semantic segmentation version of the dataset as well, and this version is used for the DMP segmentation network experiments.
The dataset is partitioned into a train, validation, and test set.
However, the semantic segmentation labels for the test set have not been made available via ground truth files or an evaluation server, so we treat the validation set as the holdout set.
Unlike social media images, satellite images can be very large, often exceeding $10,000\times10,000$ pixels, and indeed, the images in iSAID range in pixel widths from $800$ to $13,000$.
In order to reasonably fit whole large images onto GPUs for deep learning, they must be downsampled, but this introduces the issue of information being lost in the process.
Instead, following prior work~\cite{wang2022empirical, ma2021factseg, zheng2020foreground-farseg, zheng2023farseg++}, $(896\times896)$-pixel crops of the images are created with a step size of $512$, resulting in some overlap.
This results in $33,978$ training images and $11,644$ validation images.

\subsection{Evaluation Metrics}
\label{subsec:eval_metrics}
The intersection over union ($IoU$) metric is used to evaluate the performance of each model.
$$IoU = \frac{|A \cap B|}{|A \cup B|} = \frac{TP}{TP + FP + FN}$$
The average $IoU$ across each class is referred to as the mean $IoU$ ($mIoU$) and is a commonly used metric within the semantic segmentation research community:
$$mIoU = \frac{1}{C}\sum_{i=1}^CIoU_i$$
where $IoU_i$ represents the $IoU$ for the $i^{th}$ class.
The $mIoU$ provides a valuable summary of segmentation quality across all object categories present in the dataset.
While $mIoU$ is the standard for semantic segmentation research, other metrics may provide useful insights into model performance.
Therefore, the precision, recall, and F1-score for each model is also reported.
Precision measures the accuracy of the positive predictions.
It is the ratio of correctly predicted positive pixels to the total number of pixels predicted as positive:
$$Precision = \frac{TP}{TP+FP}$$
One may want to prioritize precision when there is a high cost associated with false positives.
Recall measures the proportion of actual positive pixels that are correctly predicted.
It is the ratio of correctly predicted positive pixels to the total number of actual positive pixels:
$$Recall = \frac{TP}{TP+FN}$$
Recall may be a more important metric if avoiding missed detections is a priority.
The F1-Score provides a balanced measure of precision and recall by calculating their harmonic mean:
$$F1=2 \times \frac{{Precision \times Recall}}{Precision + Recall}$$
As with $mIoU$, precision, recall, and F1-score are computed on a per-class basis and averaged, and reported as $mRecall$, $mPrecision$, and $mF1$.
Each of these reported metrics produce values between $0$ (worst possible result) and $1$ (best possible result).

\subsection{Experimental Setup}
\label{subsec:exp_setup}
Each of the semantic segmentation networks is an encoder-decoder network, and the ImageNet~\cite{deng2009imagenet} pretrained weights for the encoders provided by the original authors are utilized.
Meanwhile, the decoders are initialized randomly using Kaiming initialization~\cite{he2015delvingkaiminginit}.
For training, the hyper-parameters provided in the SegFormer and SegNeXt publications are closely followed.
The authors of EfficientViT have not released the specific training hyper-parameters used for their semantic segmentation experiments, so it also adopts the SegFormer/SegNeXt configuration in the following experiments.
Each model is trained for $160,000$ iterations, with a batch size of $16$ and using the AdamW optimizer~\cite{loshchilov2017decoupledadamw} with an initial learning rate of $0.00006$ and polynomial learning rate scheduler.
Image augmentation is applied during training to artificially increase the diversity of samples presented to each model.
Horizontal and vertical flips as well as random photometric distortions are employed, with each model's training job given the same random seed to ensure fair comparison.
At the end of training, the model is evaluated on the iSAID validation dataset without any test-time augmentation.
DMP-enabled models are first trained with the direct-in incorporation strategy.
The DMP parameters which yield a model with the highest mIoU on the iSAID validation set are then chosen for a Hybrid DMP incorporation experiment.
Therefore, each encoder has multiple direct-in experiments but only one Hybrid DMP experiment.

\subsection{Quantitative Results}
\label{subsec:quant_results}
\subsubsection{Average Results}
Table~\ref{table:isaid_dmp_val} summarizes the overall performance of each network architecture with and without DMP incorporation.
In the \textbf{Network} column, the original name refers to the baseline (e.g., SegFormer-B1), DI refers to the direct-in incorporation strategy, and HD refers to the Hybrid DMP strategy.
For each backbone (SegFormer-B1, SegNeXt-S, EfficientViT-B2) we compare the baseline to eight DMP configurations: \se sizes of original~\cite{scott2020differential_ogdmpnet}, improved~\cite{hurt2021improved}, and the two \se size groups discovered via an ECA~\cite{hurt2022evolutionary}, and for each \se size configuration, a square and disk \se shape are compared.
\begin{table}[!h]
    \centering
    \caption{iSAID DMP Model Validation Results. }
    \label{table:isaid_dmp_val}
    \resizebox{\columnwidth}{!}{%
    \begin{tabular}{l|c|c|c|c|c|c}
        \textbf{Network} & \textbf{\se Set} & \textbf{\se Shape} & \textbf{mIoU} & \textbf{mF1} & \textbf{mPrec.} & \textbf{mRec.} \\
        \hline
        \multirow{1}{*}{SegFormer-B1}         
            & -- & -- & \textbf{66.51} & \textbf{78.79} & \textbf{86.96} & \textbf{73.16} \\
        \hline
        \multirow{8}{*}{DI.SegFormer-B1}         
            & Original & Square & 62.89 & 75.57 & 85.79  & 69.06 \\
            & Original & Disk & 62.98 &	75.65 &	85.81 &	69.32 \\
            & Improved & Square & 62.60 & 75.37 & 85.44 & 69.26 \\
            & Improved & Disk & 62.64 & 75.37 & 85.71 & 68.90 \\
            & Evo. 1 & Square & 62.61 & 75.36 & 85.53 & 68.87 \\
            & Evo. 1 & Disk & 63.03 & 75.70 & 85.97 & 69.26 \\
            & Evo. 2 & Square & 62.51 & 75.23 & 85.23 & 68.91 \\
            & Evo. 2 & Disk & 63.19 & 75.85 & 85.95 & 69.70 \\
        \hline
        \multirow{1}{*}{HD.SegFormer-B1}         
            & Evo. 2 & Disk & \underline{65.84} & \underline{78.23} & \underline{86.54} & \underline{72.52} \\
        \hline
        \hline
        \multirow{1}{*}{SegNeXt-S}         
            & -- & -- & \underline{67.72} & \underline{79.59} & \textbf{85.66} & \underline{75.36} \\
        \hline
        \multirow{8}{*}{DI.SegNeXt-S}          
            & Original & Square & 65.87 & 77.88 & 85.03 & 73.03 \\
            & Original & Disk & 66.26 & 78.19 & 84.90 & 73.48 \\
            & Improved & Square & 65.54 & 77.69 & 84.10 & 73.35 \\
            & Improved & Disk & 65.94 & 77.95 & 84.41 & 73.80 \\
            & Evo. 1 & Square & 65.78 & 77.79 & 83.24 & 74.01 \\
            & Evo. 1 & Disk & 65.94 & 77.94 & 84.92 & 73.23 \\
            & Evo. 2 & Square & 66.11 & 78.11 & 84.51 & 73.71 \\
            & Evo. 2 & Disk & 65.96 & 77.95 & 84.27 & 73.58 \\
        \hline
        \multirow{1}{*}{HD.SegNeXt-S}         
            & Original & Disk & \textbf{67.97} & \textbf{79.76} & \underline{85.23} & \textbf{75.69} \\
        \hline
        \hline
        \multirow{1}{*}{EfficientViT-B2}         
            & -- & -- & \textbf{67.12} & \textbf{79.18} & \textbf{86.70} & \textbf{73.58} \\
        \hline
        \multirow{8}{*}{DI.EfficientViT-B2}     
            & Original & Square & 65.01 & 77.37 & 86.03 & 71.54 \\
            & Original & Disk & 65.47 & 77.70 & 86.57 & 71.76 \\
            & Improved & Square & 65.33 & 77.52 & 86.18 & 71.81 \\
            & Improved & Disk & 65.19 & 77.47 & 86.15 & 71.69 \\
            & Evo. 1 & Square & 65.14 & 77.46 & 85.89 & 71.80 \\
            & Evo. 1 & Disk & 64.82 & 77.11 & 84.98 & 71.77 \\
            & Evo. 2 & Square & 64.50 & 76.83 & 85.81 & 71.04 \\
            & Evo. 2 & Disk & 65.32 & 77.53 & 85.67 & 72.00 \\
        \hline
        \multirow{1}{*}{HD.EfficientViT-B2}         
            & Original & Disk & \underline{65.90} & \underline{78.22} & \underline{85.98} & \underline{72.66} \\
        \hline
        \hline
    \end{tabular}
    }
    \vspace{0.5em}
    \parbox{\columnwidth}{\footnotesize \textbf{Bolded} entries denote the best result for a metric within the results for an architecture, and \underline{underlined} entries denote the second best.}
\end{table}
Some key observations regarding the integration of the DMP into segmentation networks emerge from these results.
First, the baseline (non-DMP) models consistently outperform the models with DMP integration.
This performance gap suggests that the DMP parameters and method of integration which worked well for classification and object detection will not inherently work well for the task of semantic segmentation.
For example, SegFormer achieves a mIoU of $66.51$, while the best-performing DMP configuration only reaches $63.19$.
However, the use of the Hybrid DMP integration method results in stronger performance across all architectures compared to using the DMP features as direct input.
Notably, the Hybrid DMP SegNeXt model surpassed the baseline in the mIoU, F1, and Recall metrics, and only lags behind the baseline precision by $0.43$.
This demonstrates the potential for a Hybrid DMP approach to closely match or exceed baseline performance for semantic segmentation.
While the SegFormer and EfficientViT Hybrid DMP models did not surpass their respective baseline results, they did decrease the gap between a DMP-enabled model with each achieving the second best scores across each metric.
These results, overall, highlight the capability of DMP to improve DNN segmentation performance, given the proper configuration and feature extraction architecture.

Within the results of the DMP models, some trends emerge which highlight the importance of the selection of DMP parameters.
The use of a disk-shaped \se tends to yield better performance over a square shape.
This is particularly evident for SegFormer and SegNeXt where the improvement held for nearly all \se size configurations across nearly all metrics.
In contrast, EfficientViT only saw improvement across all metrics when using a disk-shaped \se with the original \se size configuration, and improvement in recall for the evo. 2 \se size configuration.
Furthermore, the best performing DMP SegFormer used the evo. 2 disk-shaped \se configuration, suggesting that a data-driven approach to DMP parameterization could yield better results for certain models.
Among the three base architectures, SegNeXt performs the best on all metrics, except for precision where EfficientViT achieves the highest score.
Additionally, the best DMP SegNeXt variant produces the smallest gap between a DMP-enabled architecture and the non-DMP baseline, indicating that SegNeXt may be better suited for DMP integration.
One possible explanation for this result is SegNeXt's use of large strip convolutions in its attention module which may more successfully detect the long, thin structures produced by the DMP.

\subsubsection{Per-Class Analysis}
While the mIoU score is a concise measure of performance across all classes in the validation set, analyzing the per-class IoU scores for the models may reveal important nuances which are otherwise obscured by the averaged metric.
Looking at the per-class performance can highlight the specific strengths and weaknesses of each model and identify which object categories are helped or hindered by the use of the DMP.
The following analysis focuses on the per-class performance of the baseline, best performing direct-in, DMP-enabled network, and the Hybrid DMP network for each encoder.
Tables \ref{tab:perclass_segformer}, \ref{tab:perclass_segnext}, and ~\ref{tab:perclass_efficientvit} display the per-class IoU scores for each encoder.
%
%
\begin{table}[!t]
    \centering
    \caption{iSAID Per-Class IoU — SegFormer-B1}
    \label{tab:perclass_segformer}
    \resizebox{\columnwidth}{!}{%
    \small                    
    \begin{tabular}{l|c|c|c}
        \textbf{Class} & \textbf{Baseline} & \textbf{Direct-In DMP} & \textbf{Hybrid DMP} \\
        \hline
        Background            & \textbf{99.10} & 99.02 & \textbf{99.10} \\
        \hline
        Ship                  & \textbf{70.74} & 67.90 & \underline{70.60} \\
        \hline
        Storage Tank          & \textbf{75.31} & 74.64 & \underline{74.90} \\
        \hline
        Baseball Diamond      & \textbf{73.19} & 67.47 & \underline{72.59} \\
        \hline
        Tennis Court          & \textbf{88.52} & 87.75 & \underline{88.04} \\
        \hline
        Basketball Court      & \textbf{66.27} & 63.97 & \underline{64.04} \\
        \hline
        Ground Track Field  & \textbf{58.79} & 48.50 & \underline{57.95} \\
        \hline
        Bridge                & \underline{39.42} & \textbf{40.14} & 37.55 \\
        \hline
        Large Vehicle         & \underline{65.56} & 65.13 & \textbf{65.93} \\
        \hline
        Small Vehicle         & \underline{52.11} & 48.90 & \textbf{52.63} \\
        \hline
        Helicopter            & \textbf{44.56} & 38.73 & \underline{41.33} \\
        \hline
        Swimming Pool         & \underline{50.88} & 28.00 & \textbf{51.47} \\
        \hline
        Roundabout            & \underline{60.51} & \textbf{63.66} & 59.13 \\
        \hline
        Soccer Ball Field     & \textbf{77.43} & 74.84 & \underline{76.36} \\
        \hline
        Plane                 & \textbf{85.89} & 83.23 & \underline{85.75} \\
        \hline
        Harbor                & 55.81 & \textbf{56.53} & \underline{56.03} \\
        \hline
    \end{tabular}
    }
    \vspace{0.5em}
    \parbox{\columnwidth}{\footnotesize \textbf{Bolded} entries denote the best result for a metric within the results for an architecture, and \underline{underlined} entries denote the second best.}
\end{table}

%
\begin{table}[!t]
    \centering
    \caption{iSAID Per-Class IoU — SegNeXt-S}
    \label{tab:perclass_segnext}
    \resizebox{\columnwidth}{!}{%
    \small                    
    \begin{tabular}{l|c|c|c}
        \textbf{Class} & \textbf{Baseline} & \textbf{Direct-In DMP} & \textbf{Hybrid DMP} \\
        \hline
        Background            & \underline{99.15} & 99.11 & \textbf{99.16} \\
        \hline
        Ship                  & \underline{71.26} & 69.07 & \textbf{71.67}\\
        \hline
        Storage Tank          & 75.17 & \underline{75.33} & \textbf{75.95} \\
        \hline
        Baseball Diamond      & \textbf{79.89} & 75.03 & \underline{78.16 }\\
        \hline
        Tennis Court          & 88.47 & \underline{88.92} & \textbf{89.08} \\
        \hline
        Basketball Court      & \underline{68.80} & \textbf{69.77} & 66.77 \\
        \hline
        Ground Track Field  & \textbf{61.78} & 57.98 & \underline{61.09} \\
        \hline
        Bridge                & \textbf{39.21} & \underline{38.41} & 37.48 \\
        \hline
        Large Vehicle         & \textbf{66.00} & 65.44 & \underline{65.74} \\
        \hline
        Small Vehicle         & \underline{51.70} & 50.07 & \textbf{51.87} \\
        \hline
        Helicopter            & \textbf{44.57} & \underline{43.62} & 42.28 \\
        \hline
        Swimming Pool         & \underline{45.62} & 31.15 & \textbf{51.65} \\
        \hline
        Roundabout            & 68.69 & \textbf{77.03} & \underline{72.09 }\\
        \hline
        Soccer Ball Field     & \underline{78.25} & 77.95 & \textbf{79.72} \\
        \hline
        Plane                 & \textbf{85.58} & 83.95 & \underline{85.52} \\
        \hline
        Harbor                & \textbf{59.35} & 57.38 & \underline{59.29} \\
        \hline
    \end{tabular}
    }
    \vspace{0.5em}
    \parbox{\columnwidth}{\footnotesize \textbf{Bolded} entries denote the best result for a metric within the results for an architecture, and \underline{underlined} entries denote the second best.}
\end{table}
%

%
\begin{table}[!t]
    \centering
    \caption{iSAID Per-Class IoU — EfficientViT-B2}
    \label{tab:perclass_efficientvit}
    \resizebox{\columnwidth}{!}{%
    \small                    
    \begin{tabular}{l|c|c|c}
        \textbf{Class} & \textbf{Baseline} & \textbf{Direct-In DMP} & \textbf{Hybrid DMP} \\
        \hline
        Background            & \textbf{99.11 }& 99.07 & \underline{99.10} \\
        \hline
        Ship                  & \textbf{70.56} & 69.40 & \underline{70.22} \\
        \hline
        Storage Tank          & \textbf{74.78} & \underline{74.47} & 73.85 \\
        \hline
        Baseball Diamond      & \textbf{76.12} & \underline{75.91} & 74.31 \\
        \hline
        Tennis Court          & \textbf{89.18} & 87.34 & \underline{88.57} \\
        \hline
        Basketball Court      & \underline{66.41} & \textbf{67.93} & 62.33 \\
        \hline
        Ground Track Field  & \underline{58.74} & 54.02 & \textbf{58.98} \\
        \hline
        Bridge                & \underline{42.40} & \textbf{42.48} & 37.84 \\
        \hline
        Large Vehicle         & \underline{65.49} & 64.91 & \textbf{65.63} \\
        \hline
        Small Vehicle         & \textbf{51.39} & 49.83 & \underline{50.90} \\
        \hline
        Helicopter            & 38.60 & \textbf{41.26} & \underline{40.58} \\
        \hline
        Swimming Pool         & \textbf{51.09} & 32.12 & \underline{48.26} \\
        \hline
        Roundabout            & \underline{67.61} & \textbf{68.21} & 63.41 \\
        \hline
        Soccer Ball Field     & \textbf{78.60} & \underline{78.41} & 78.33 \\
        \hline
        Plane                 & \underline{85.32} & 84.16 & \textbf{85.55} \\
        \hline
        Harbor                & \textbf{58.48} & \underline{57.99} & 56.55 \\
        \hline
    \end{tabular}
    }
    \vspace{0.5em}
    \parbox{\columnwidth}{\footnotesize \textbf{Bolded} entries denote the best result for a metric within the results for an architecture, and \underline{underlined} entries denote the second best.}
\end{table}
A consistent trend across all three encoders is the performance of the direct-input DMP model.
While it generally results in the lowest mIoU, it also achieves the highest IoU for a specific subset of classes characterized by strong, simple geometric shapes.
Notably, the Direct-Input model records the best performance for \textit{roundabout} (SegFormer, SegNeXt, EfficientViT), \textit{Bridge} (SegFormer, EfficientViT), and \textit{basketball court} (SegNeXt, EfficientViT).
One possible interpretation of these results is that these classes are largely defined by their shape and structure, where color and texture are less informative or can be misleading.
The direct-input DMP models are forced to learn from the grayscale and shape information, potentially allowing it to achieve superior performance on objects who are most distinct in their shape.
However, the poor performance of the direct-input DMP models on most other classes demonstrates the importance of color information for accurate segmentation, highlighting a significant drawback of this method.
The Hybrid DMP models are often able to mitigate the weaknesses of the direct-input approach, demonstrating the value of fusing both color-based and DMP-based features.
In particular, the Hybrid DMP approach shows success for the SegNeXt-S encoder, where it achieved the highest IoU in 7 out of 16 classes.
The classes that benefit most from the Hybrid DMP approach appear to be those which are both distinct in shape and color.
For example, the \textit{swimming pool} class sees a more pronounced performance increase when using the Hybrid DMP model in both the SegFormer and SegNeXt Hybrid DMP models.
This suggests that the model is effectively able to utilize the shape information from the DMP stream, likely from the oblong or curvilinear appearance of pools in overhead imagery, as well as their distinct blue color from the RGB stream.
A similar pattern is observed for the \textit{large vehicle} and \textit{small vehicle} classes which often have distinct shapes from their overall structure and sub-components like windows as well as their color contrasting with the surroundings.
The baseline models, particularly SegFormer-B1 and EfficientViT-B2, demonstrate their strength on classes that are less defined by a simple shape and more by complex internal patterns, textures, and color variations.
The baseline models consistently perform best on classes such as \textit{baseball diamond}, \textit{ground track field}, \textit{helicopter}, and \textit{plane}.
These objects often contain details that may be better captured by the original RGB image such as the distinct colors of a baseball infield and outfield or colored markings on aircraft like camouflage.
The per-class analysis suggests that there is no one-size-fits-all solution for integrating DMP features.
The direct-input method, while demonstrating inferior performance overall, can be uniquely effective for mostly geometric targets.
Alternatively, the Hybrid DMP approach can achieve superior performance by combining explicit shape information with rich color features.

\subsection{Qualitative Results}
\label{subsec:qual_results}
\begin{figure*}
    \centering
    \subfloat[Input Image]{
        \includegraphics[width=0.22\textwidth]{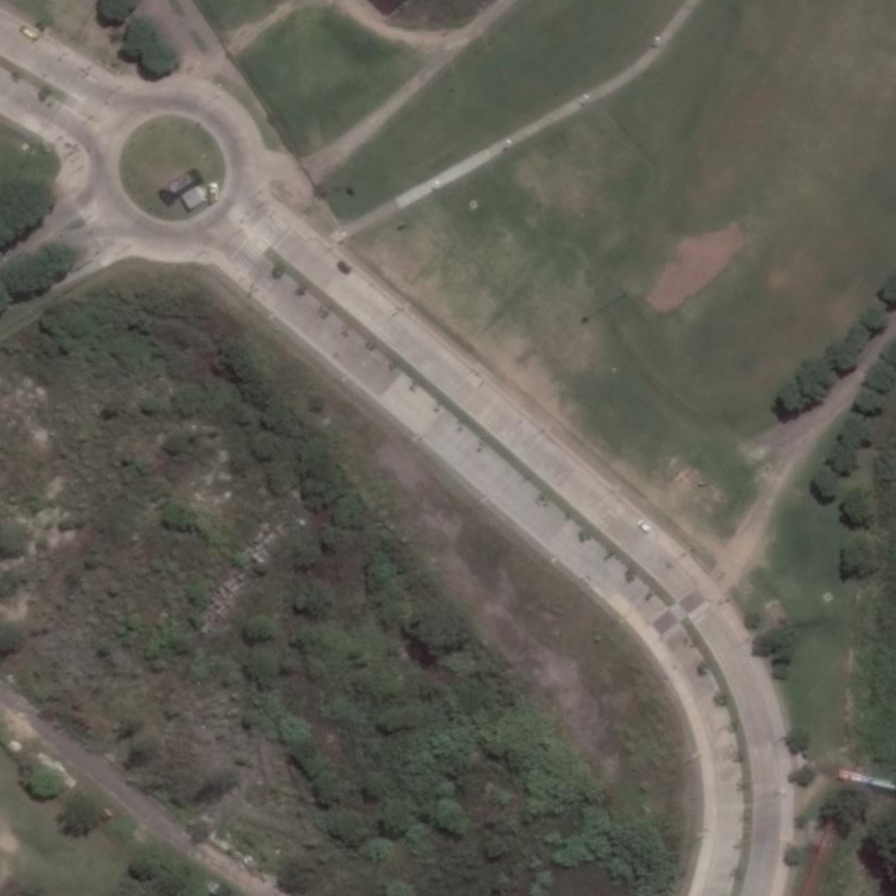}
    } \hfill
    \subfloat[Baseline]{
        \includegraphics[width=0.22\textwidth]{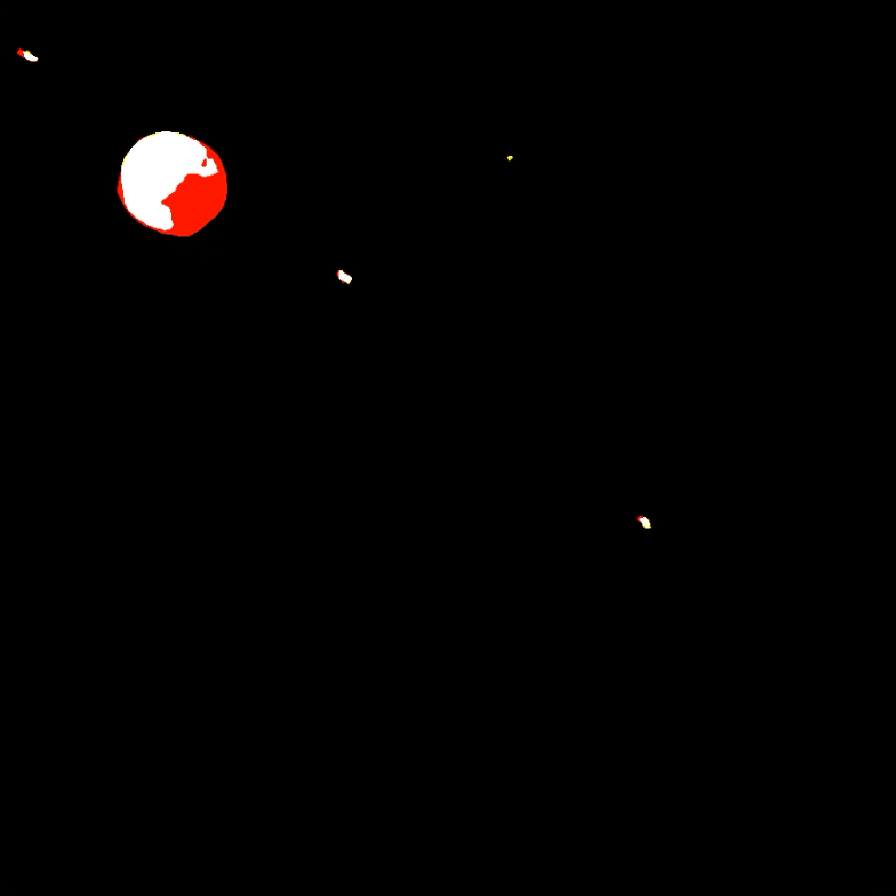}
    } \hfill
    \subfloat[Direct-In]{
        \includegraphics[width=0.22\textwidth]{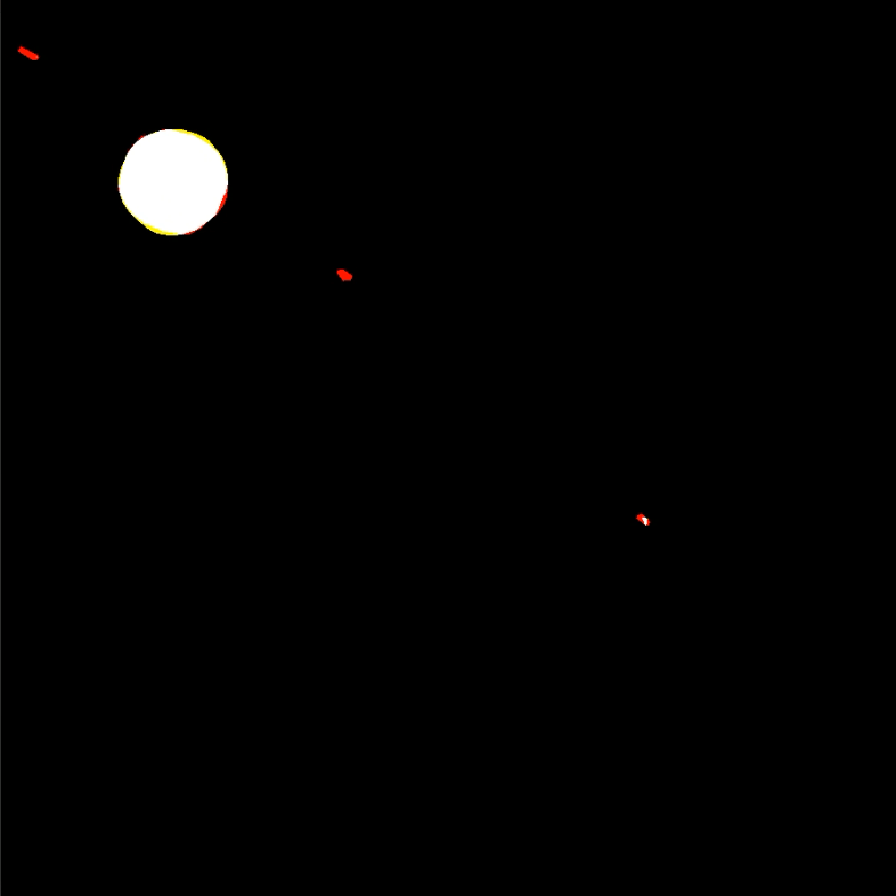}
    } \hfill
    \subfloat[Hybrid DMP]{
        \includegraphics[width=0.22\textwidth]{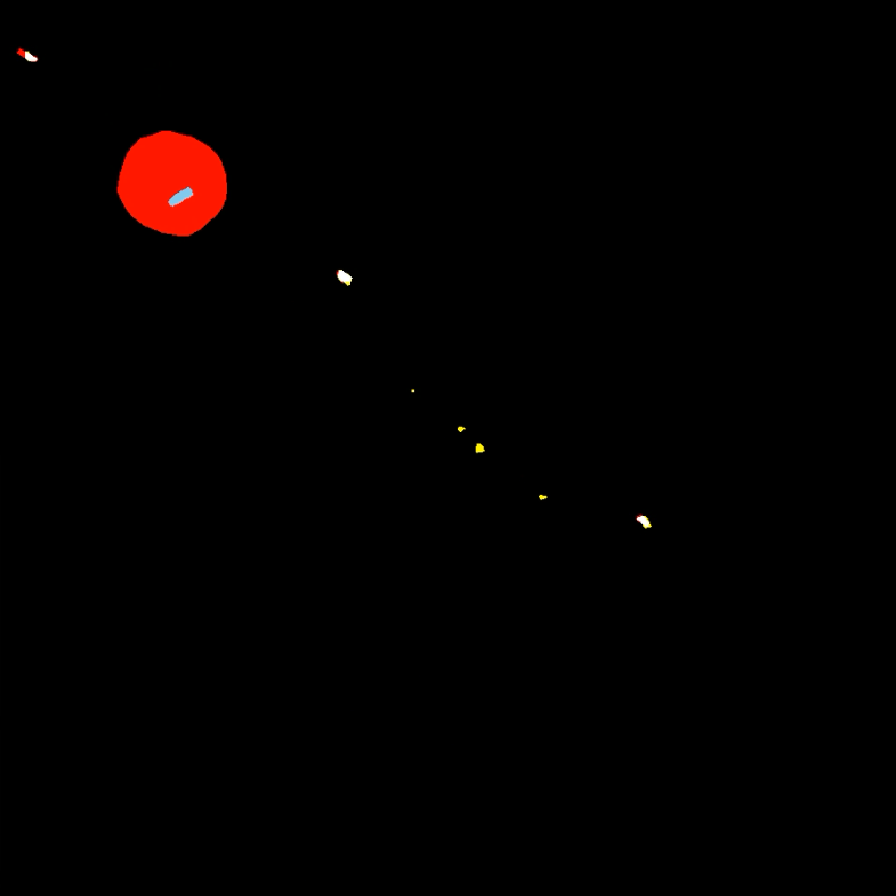}
    }
    \caption{EfficientViT model performance on the \textit{roundabout} class. Note how the Direct-In method achieves the strongest segmentation (white = \textbf{true positive}, yellow = \textbf{false positive}, red = \textbf{false negative}, and blue = \textbf{true positive, but wrong foreground class}).}
    \label{efficientvit_roundabout}
\end{figure*}

To complement the quantitative metrics presented in Section~\ref{subsec:quant_results}, this section provides a qualitative analysis of the segmentation results.
By visually inspecting the output masks from the baseline, best-performing direct-input DMP, and Hybrid DMP models, we can gain a more nuanced understanding of their respective strengths and weaknesses.
The following figures illustrate performance on representative images from the iSAID validation set.
Note that in the visualizations, white = \textbf{true positive}, yellow = \textbf{false positive}, red = \textbf{false negative}, and blue = \textbf{true positive, but wrong foreground class}.
Figure~\ref{efficientvit_roundabout} demonstrates a scenario where the direct-input DMP model excels.
The selected image prominently features a roundabout, a class for which the direct-input models achieved the highest IoU across all three architectures.
While the baseline model is able to identify some pixels in the roundabout, it struggles to segment out the entire object as much as the direct-in model.
This example provides support for the hypothesis that for classes which are more defined by their structure than their texture or spectral appearance, the shape information provided by the DMP can be beneficial.
For this specific case, the interior of the roundabout is difficult to distinguish from the grassy background area within the input image, for example.
Interestingly, the Hybrid DMP model performs worse than both the baseline and the direct-input model suggesting that the current method of feature fusion does not necessarily always take advantage of the strengths of each modality.
%

\begin{figure*}
    \centering
    \subfloat[Input Image]{
        \includegraphics[width=0.22\textwidth]{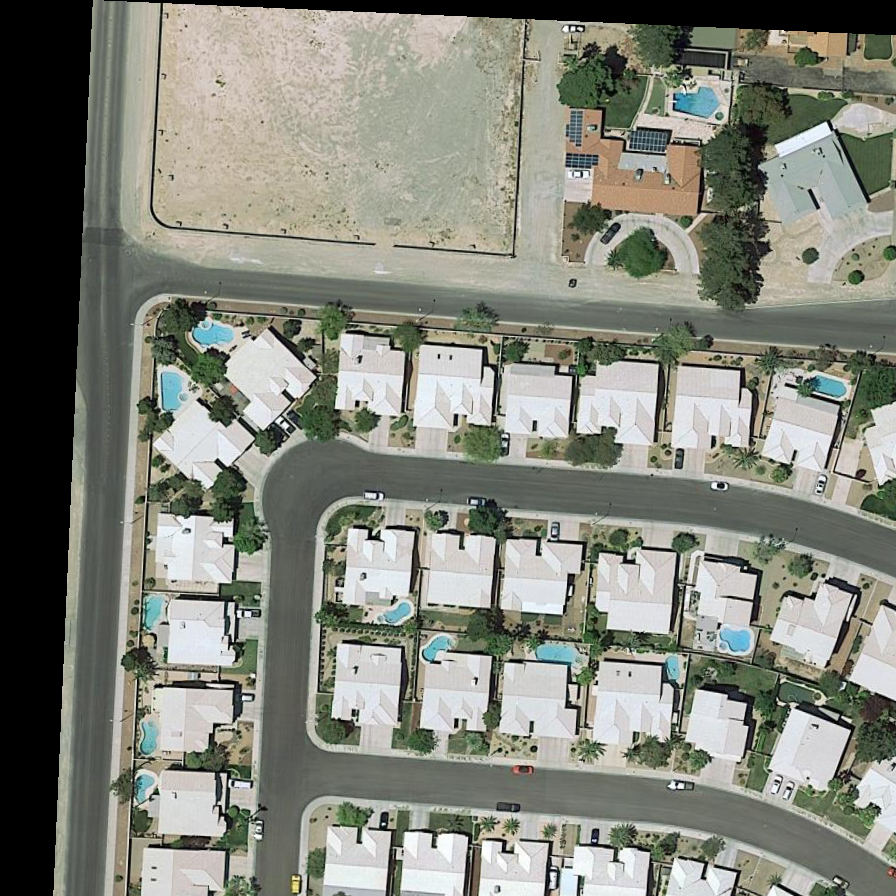}
    } \hfill
    \subfloat[Baseline]{
        \includegraphics[width=0.22\textwidth]{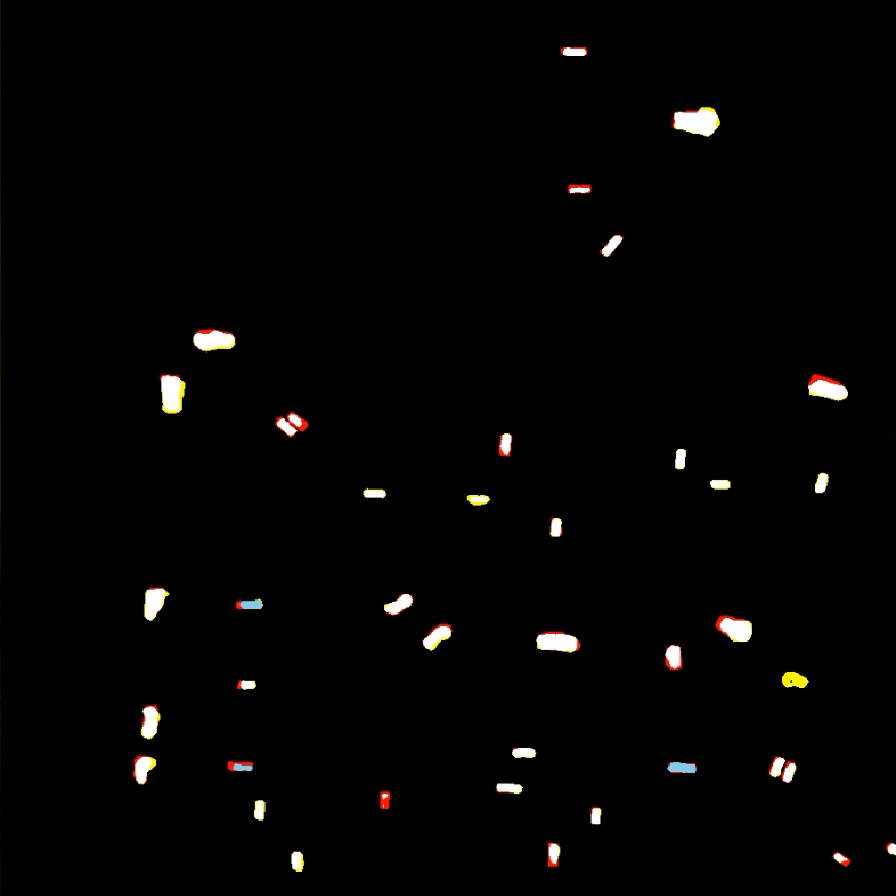}
    } \hfill
    \subfloat[Direct-In]{
        \includegraphics[width=0.22\textwidth]{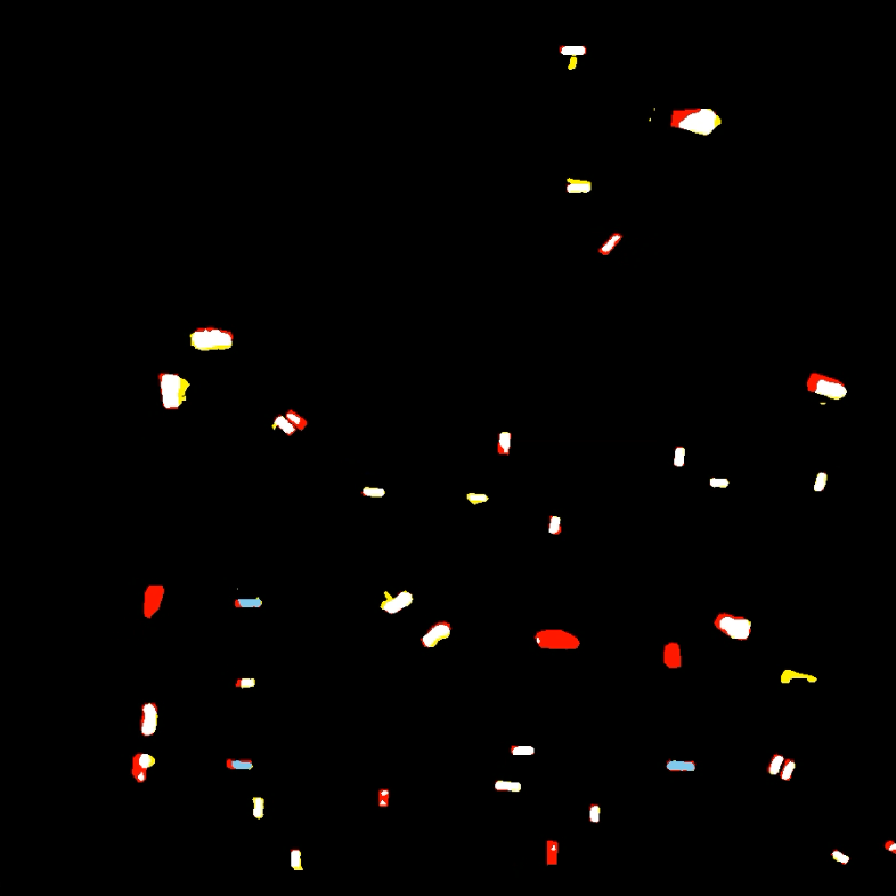}
    } \hfill
    \subfloat[Hybrid DMP]{
        \includegraphics[width=0.22\textwidth]{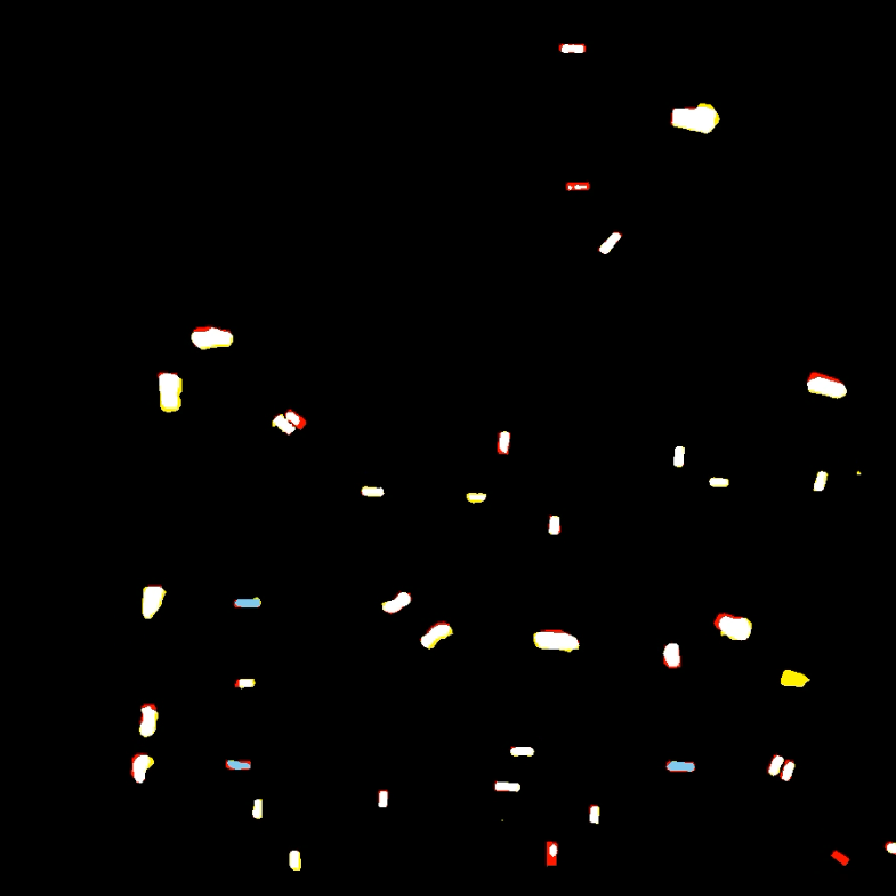}
    } \hfill
    \caption{SegFormer model performance on the \textit{swimming pool} class. Note how some pools are missed only by the Direct-In method (white = \textbf{true positive}, yellow = \textbf{false positive}, red = \textbf{false negative}, and blue = \textbf{true positive, but wrong foreground class}).}
    \label{segformer_swimming_pool}
\end{figure*}

%
Figure~\ref{segformer_swimming_pool} highlights the strength of the Hybrid DMP approach on classes where both shape and color are critical identifiers, such as in the case of the \textit{swimming pool} class.
The direct-input models consistently fail to segment a number of the swimming pools in the image.
In contrast, the baseline and Hybrid DMP models are able to extract the pools missed by the direct-in models.
However, compared to the baseline models, the Hybrid DMP models tend to slightly decrease the number of incorrect pixel classifications near the edges of the pools.
This suggests that the Hybrid DMP is capable of utilizing both the shape and color information provided by each stream to produce segmentations more accurate than each individual method alone.
%

\begin{figure*}
    \centering
    \subfloat[Input Image]{
        \includegraphics[width=0.22\textwidth]{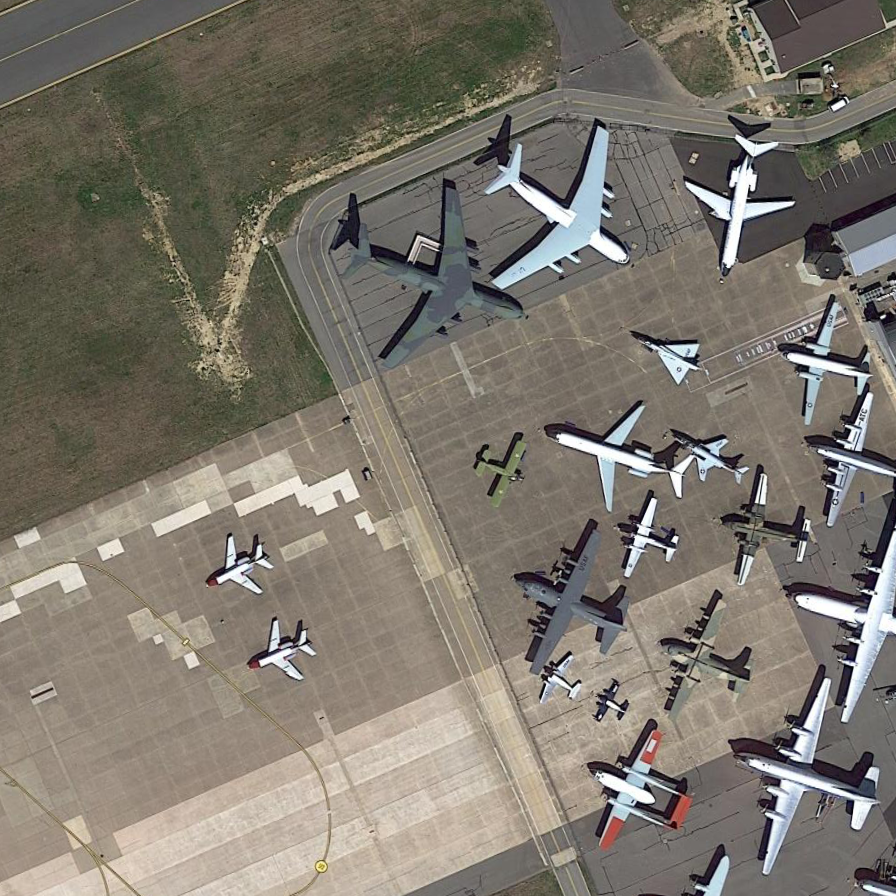}
    } \hfill
    \subfloat[Baseline]{
        \includegraphics[width=0.22\textwidth]{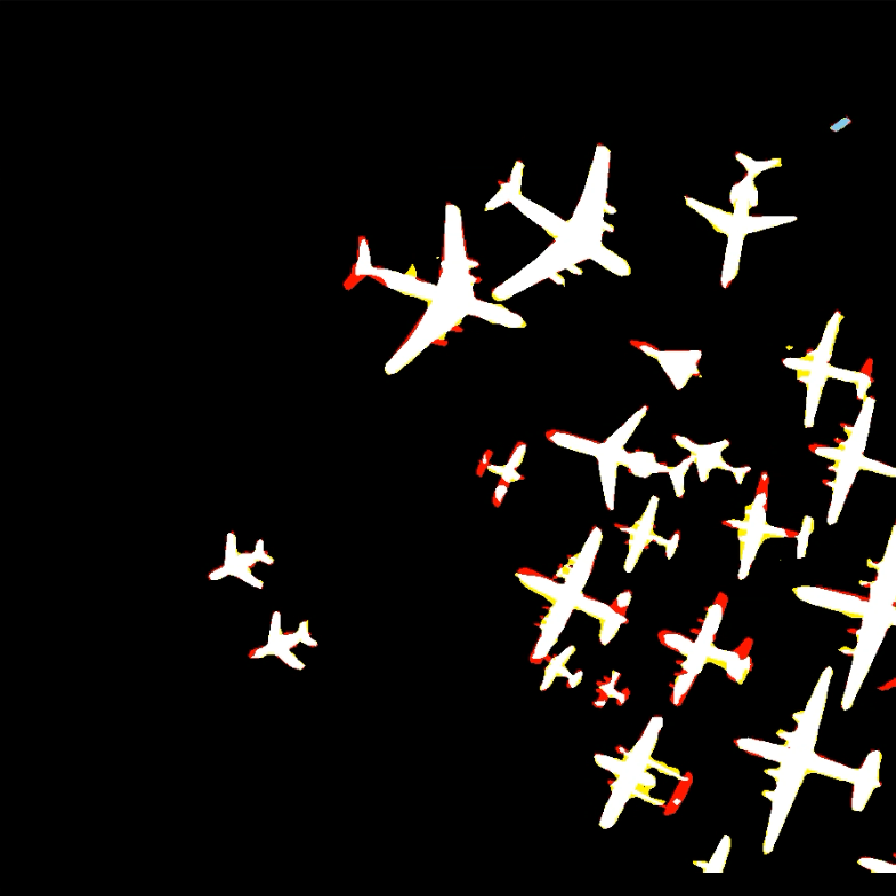}
    } \hfill
    \subfloat[Direct-In]{
        \includegraphics[width=0.22\textwidth]{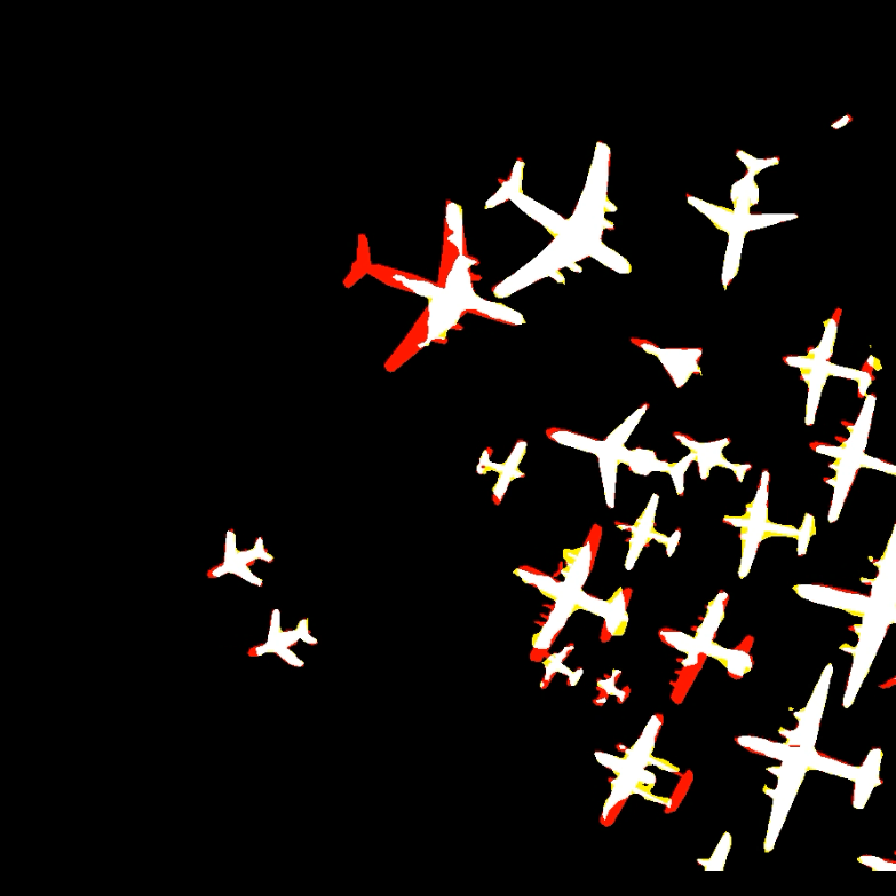}
    } \hfill
    \subfloat[Hybrid DMP]{
        \includegraphics[width=0.22\textwidth]{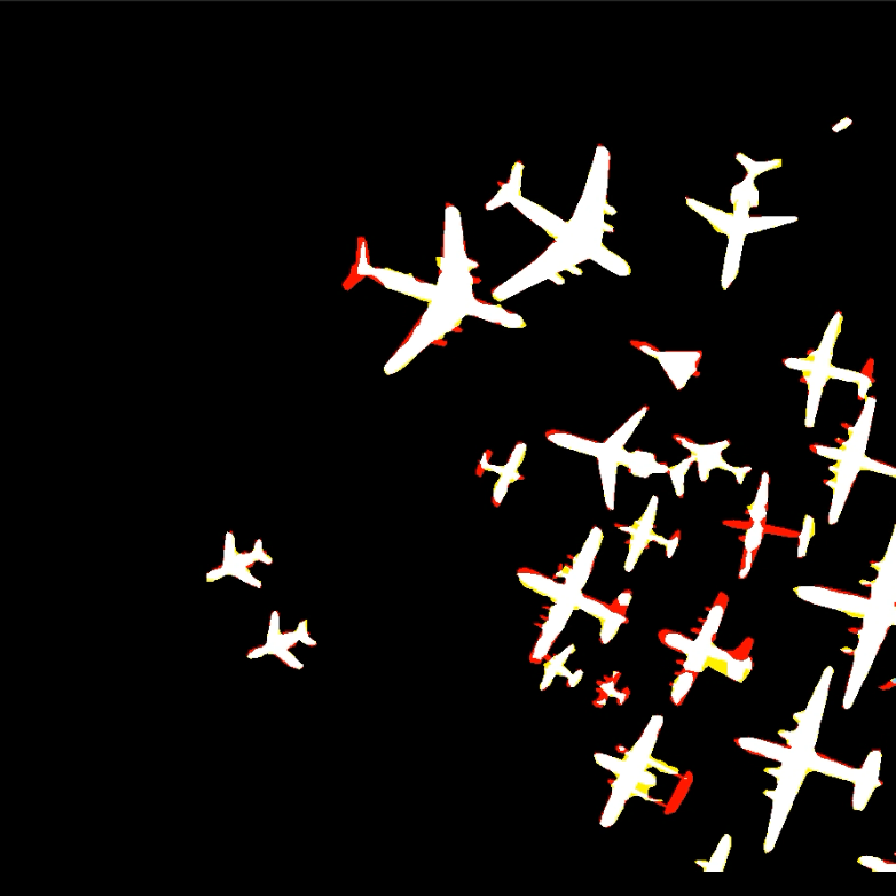}
    } \hfill
    \caption{SegNeXt model performance on the \textit{plane} class. Note the Direct-In model's failure to segment the large, camouflaged plane (white = \textbf{true positive}, yellow = \textbf{false positive}, red = \textbf{false negative}, and blue = \textbf{true positive, but wrong foreground class}).}
    \label{segnext_plane}
\end{figure*}

%
Figure~\ref{segnext_plane} demonstrates a distinct failure mode of the direct-in method.
Note how the largest, camouflaged aircraft is poorly segmented by the direct-in model.
Conversely, the baseline and Hybrid DMP models are able to much more accurately segment the camouflaged plane.
This highlights the importance of color information, which the direct-input method does not have access to.
Similarly, in the two smaller aircraft towards the bottom left region of the image, the direct-in model misclassifies the red colored nose.
%

\section{Conclusion}
\label{sec:conclusion}
In this work, we extended DMPNets to semantic segmentation in remote sensing imagery.
Our experiments revealed that existing methods for DMPNet classification and object detection do not directly translate to superior performance in this task.
The Hybrid DMP approach, however, which fuses features from parallel RGB and DMP encoders, proved more effective, matching and even surpassing baseline performance with the SegNeXt architecture.
This result demonstrates the potential for the DMP's multi-scale shape extraction to provide value to deep neural networks.
However, these results also suggest that the DMP features must complement, rather than replace, the spectral information present in the original RGB image.
Based on these findings, avenues for further research are proposed.
First, evolutionary computation algorithms could be applied to search for segmentation-specific DMP differentials, with methods such as proxy fitness functions potentially reducing search time.
Second, strategies for improving the efficiency of DMP computation, particularly for large structuring element sizes, such as downsampling the input image prior to computation or using iterative applications of the elementary structuring element could be explored.
Finally, the Hybrid DMP architecture could be refined by employing a lighter-weight encoder for the DMP stream, which may better exploit the sparse nature of DMP features, and by incorporating more sophisticated feature fusion mechanisms such as attention.

\bibliographystyle{IEEEtran}
\bibliography{refs}

\vfill

\end{document}